\pdfoutput=1
\documentclass{article}

\PassOptionsToPackage{numbers, compress}{natbib}
\newtheorem{theorem}{Theorem}

\usepackage[final]{neurips_data_2022}

\usepackage[utf8]{inputenc} 
\usepackage[T1]{fontenc}    
\usepackage{hyperref}       
\usepackage{url}            
\usepackage{booktabs}       
\usepackage{amsfonts}       
\usepackage{nicefrac}       
\usepackage{bm}
\usepackage{microtype}      
\usepackage{xcolor}         
\usepackage[ruled,vlined]{algorithm2e}    
\usepackage{algpseudocode}
\usepackage{amssymb}
\usepackage{pifont}
\usepackage{graphicx}
\usepackage{floatrow}
\usepackage{authblk}
\newfloatcommand{capbtabbox}{table}[][\FBwidth]
\usepackage[export]{adjustbox}
\usepackage[framemethod=tikz]{mdframed}

\usepackage{float}
\floatstyle{plaintop}
\restylefloat{table}
\usepackage{multirow}
\usepackage{wrapfig}

\usepackage[T1]{fontenc}
\usepackage[utf8]{inputenc}
\usepackage{babel}
\usepackage[font=small,labelfont=bf]{caption}

\definecolor{darkgreen}{rgb}{0,0.7,0.5}
\newcommand{\cmark}{\ding{51}}%
\newcommand{\xmark}{\ding{55}}%
 
\newcommand{\xhdr}[1]{\vspace{0em}\noindent{{\bf #1.}}}
\newcommand{\ie}{\textit{i.e., \xspace}}
\newcommand{\eg}{\textit{e.g., \xspace}}

\newcommand{\std}[1]{\scriptsize{$\pm$#1}}

\newcommand{\name}{OpenXAI\xspace}

\usepackage{colortbl}

\definecolor{Gray}{gray}{0.9}
\definecolor{LightCyan}{rgb}{0.88,1,1}
\newcommand{\hide}[1]{}

\newcommand{\LIME}{LIME\xspace}
\newcommand{\SHAP}{SHAP\xspace}

\newcommand{\Grads}{Vanilla Gradients\xspace}
\newcommand{\SmoothGrad}{SmoothGrad\xspace}
\newcommand{\IntGrad}{Integrated Gradients\xspace}
\newcommand{\GradtimesInput}{Gradient x Input\xspace}

\usepackage{array,ragged2e}
\newcolumntype{P}[1]{>{\RaggedRight\arraybackslash}p{#1}}

\title{\name: Towards a Transparent Evaluation of \\Post hoc Model Explanations}

\author[1]{Chirag Agarwal}
\author[1]{Dan Ley}
\author[1]{Satyapriya Krishna}
\author[1]{Eshika Saxena*}
\author[1]{Martin Pawelczyk}
\author[4]{Nari Johnson}
\author[1]{Isha Puri*}
\author[1]{Marinka Zitnik}
\author[1]{Himabindu Lakkaraju}
\affil[1]{Harvard University}
\affil[4]{Carnegie Mellon University}

\begin{document}

\maketitle
\def\thefootnote{*}\footnotetext{Work done by authors when they were at Harvard University.}
\begin{abstract}
While several types of post hoc explanation methods have been proposed in recent literature, there is very little work on systematically benchmarking these methods. Here, we introduce \name, a comprehensive and extensible open-source framework for evaluating and benchmarking post hoc explanation methods. \name comprises of the following key components: (i) a flexible synthetic data generator and a collection of diverse real-world datasets, pre-trained models, and state-of-the-art feature attribution methods, and (ii) open-source implementations of eleven quantitative metrics for evaluating faithfulness, stability (robustness), and fairness of explanation methods, 
in turn providing comparisons of several explanation methods across a wide variety of metrics, models, and datasets.
\name is easily extensible, as users can readily evaluate custom explanation methods and incorporate them into our leaderboards. Overall, \name provides an automated end-to-end pipeline that not only simplifies and standardizes the evaluation of post hoc explanation methods, but also promotes transparency and reproducibility in benchmarking these methods. While the first release of OpenXAI supports only tabular datasets, the explanation methods and metrics that we consider are general enough to be applicable to other data modalities. \name datasets and models, implementations of state-of-the-art explanation methods and evaluation metrics, are publicly available at \href{https://github.com/AI4LIFE-GROUP/OpenXAI}{this GitHub link}.

\end{abstract}

\section{Introduction}
\label{sec:intro}
As predictive models are increasingly deployed in critical domains (e.g., healthcare, law, and finance), there has been a growing emphasis on explaining the predictions of these models to decision-makers (\eg doctors, and judges) so that they can understand the rationale behind model predictions, and determine if and when to rely on these predictions.
To this end, various techniques have been proposed in recent literature to generate post hoc explanations of individual predictions made by complex ML models.
Several of such \emph{local explanation methods} output the influence of each of the features on the model's prediction, and are therefore referred to as \emph{local feature attribution methods}.
\hide{These methods can be further categorized into \textit{gradient-based} or \textit{perturbation-based} methods depending on their approach to estimating each feature's influence on the prediction.
Popular state-of-the-art perturbation-based methods include \LIME~\cite{ribeiro16:kdd} and \SHAP~\cite{lundberg2017unified}, and gradient-based ones include \Grads~\cite{simonyan2013saliency}, \GradtimesInput~\cite{shrikumar2017learning}, \SmoothGrad~\cite{smilkov2017smoothgrad}, and \IntGrad~\cite{sundararajan2017axiomatic}.}
Due to their generality, feature attribution methods are increasingly being utilized to explain complex models in medicine, finance, law, and science~\cite{elshawi2019interpretability,ibrahim2019global,whitmore2016mapping}. 
Thus, it is critical to ensure that the explanations generated by these methods are \emph{reliable} so that relevant stakeholders and decision makers are provided with credible information about the underlying models~\cite{aivodji2019fairwashing}.

\looseness=-1
Prior works have studied several notions of explanation reliability such as \emph{faithfulness} (or fidelity)~\cite{zhou2021evaluating,liu2021synthetic,hooker2018evaluating}, \emph{stability} (or robustness)~\cite{alvarez2018robustness,agarwal2022rethinking}, and \emph{fairness}~\cite{dai2022fairness,balagopalan2022road}, and proposed metrics for quantifying these notions. Many of these works also demonstrated through small-scale experiments or qualitative analysis that certain explanation methods are not effective w.r.t. specific notions of reliability. For instance,~\citet{alvarez2018robustness} visualized the explanations generated by some of the popular gradient based explanation methods ~\cite{simonyan2013saliency,shrikumar2017learning,smilkov2017smoothgrad,sundararajan2017axiomatic} for MNIST images, and showed that they are not robust to small input perturbations. However, it is unclear if such findings generalize beyond the settings studied. More broadly, one of the biggest open questions which has far-reaching implications for the progress of explainable AI (XAI) research is: \emph{which explanation methods} are effective w.r.t. \emph{which notions of reliability} and \emph{under what conditions}? ~\cite{krishna2022disagreement}. A first step towards answering this question involves systematically benchmarking explanation methods in a reproducible and transparent manner. However, the increasing diversity of explanation methods, and the plethora of evaluation settings and metrics outlined in existing research without standardized open-source implementations make it rather challenging to carry out such benchmarking efforts. 

In this work, we address the aforementioned challenges by introducing OpenXAI, a comprehensive and extensible open-source framework for systematically and efficiently benchmarking explanation methods in a transparent and reproducible fashion. More specifically, our work makes the following key contributions: 

\begin{enumerate}
    \item We introduce the OpenXAI framework, an \emph{open-source ecosystem} designed to support systematic, reproducible, and efficient evaluations of post hoc explanation methods.  OpenXAI unifies the existing scattered repositories of datasets, models, and evaluation metrics, and provides a \emph{simple and easy-to-use API} that enables researchers and practitioners to benchmark explanation methods using just a few lines of code (Section~\ref{sec:overview}).
    \item Our OpenXAI framework currently provides open-source implementations and ready-to-use API interfaces for \emph{six state-of-the-art feature attribution methods} (\LIME, \SHAP, \Grads, \GradtimesInput, \SmoothGrad, and \IntGrad), and \emph{eleven quantitative metrics} to evaluate the faithfulness, stability, and fairness of feature attribution methods. In addition, it includes a comprehensive collection of \emph{seven real-world datasets} spanning diverse real-world domains, and \emph{sixteen different pre-trained models}. \name also introduces \emph{a novel and flexible synthetic data generator} to synthesize datasets of varying sizes, complexity, and dimensionality which facilitate the construction of reliable ground truth explanations (Section~\ref{sec:overview}).
   \item The \name framework is \emph{easily extensible} i.e., researchers and practitioners can readily incorporate custom explanation methods, datasets, predictive models, and evaluation metrics into our framework and benchmarks (Section~\ref{sec:overview}).
   \item Lastly, using \name, we \emph{perform rigorous empirical benchmarking} of the aforementioned state-of-the-art feature attribution methods to determine which methods are effective w.r.t. what notions of reliability across a wide variety of datasets and predictive models (Section~\ref{sec:expt}). 
\end{enumerate}

Overall, our OpenXAI framework provides an end-to-end pipeline that unifies, simplifies, and standardizes several existing workflows to evaluate explanation methods. By enabling systematic and efficient evaluation and benchmarking of existing and new explanation methods, our OpenXAI framework can inform and accelerate new research in the emerging field of XAI. \name will be regularly updated and welcomes input from the community.

\color{black}
\xhdr{Related Work} Our work builds on the vast literature in explainable AI. Here, we discuss closely related works and their connections to our benchmark (see the Appendix for further discussion).

\emph{Evaluation Metrics for Post hoc Explanations}: Prior research has studied several notions of explanation reliability, namely, \emph{faithfulness} (or fidelity), \emph{stability} (or robustness), and \emph{fairness}~\cite{liu2021synthetic,zhou2021evaluating,alvarez2018robustness,dai2022fairness}. While the faithfulness notion captures how faithfully a given explanation captures the true behavior of 
the underlying model~\cite{zhou2021evaluating,liu2021synthetic,hooker2018evaluating}, stability ensures that explanations do not change drastically with small perturbations to the input~\cite{ghorbani2019interpretation,alvarez2018robustness}. The fairness notion, on the other hand, ensures that there are no group-based disparities in the faithfulness or stability of explanations~\cite{dai2022fairness}. To this end,
prior works~\cite{liu2021synthetic,slack2021reliable,zhou2021evaluating,alvarez2018robustness,dai2022fairness,hooker2018evaluating} proposed various evaluation metrics to quantify the aforementioned notions. For instance,~\citet{petsiuk2018rise} measured the change in the probability of the predicted class when important features (as identified by an explanation) are deleted from or introduced into the data instance. A sharp change in the probability implies a high degree of explanation faithfulness.~\citet{alvarez2018robustness} loosely quantified stability as the maximum change in the resulting explanations when small perturbations are made to a given instance.~\citet{dai2022fairness} quantified unfairness of explanations as the difference between the faithfulness (or stability) metric values averaged over instances in the majority and the minority subgroups.

\emph{XAI Libraries and Benchmarks: } 
Prior works have introduced XAI libraries and benchmarks, the most popular among them being Captum~\citep{kokhlikyan2020captum}, Quantus~\citep{hedstrom2022quantus}, XAI-Bench~\citep{liu2021synthetic}, and \href{https://shap.readthedocs.io/en/latest/index.html}{SHAP Benchmark}. Below, we provide a brief description for each, and detail how our work differs from them. 

While \textbf{\emph{Captum library}}~\citep{kokhlikyan2020captum} is an open-source library that provides implementations and APIs for various state-of-the-art explanation methods, its focus is not on evaluating and/or benchmarking these methods which is the main goal of our work.
\textbf{\emph{Quantus library}}~\cite{hedstrom2022quantus}, on the other hand, provides implementations of certain evaluation metrics to measure the faithfulness and stability/robustness of explanation methods. However, it does not focus on benchmarking explanation methods or providing public dashboards to compare the performance of these methods. Furthermore, the stability/robustness measures~\citep{alvarez2018robustness} supported by Quantus are somewhat outdated and have been superseded by recently proposed metrics~\citep{agarwal2022rethinking}. In addition, Quantus does not support any fairness metrics to evaluate disparities in the quality of explanations which  is very important in real-world settings such as healthcare, criminal justice, and policy. In contrast, OpenXAI not only subsumes popular faithfulness and stability/robustness metrics supported by Quantus but also supports 7 new metrics to measure the faithfulness, stability/robustness, as well as the fairness of explanation methods~\citep{agarwal2022rethinking,dai2022fairness,krishna2022disagreement}. In addition, OpenXAI focuses on systematically benchmarking state-of-the-art explanation methods and providing public dashboards to readily compare these methods. 

\textbf{\emph{SHAP benchmark}}~\cite{shapbench} only focuses on evaluating and comparing different variants of SHAP~\citep{lundberg17:a-unified} via certain faithfulness metrics which are similar to the Prediction Gap on Important (PGI) and Unimportant (PGU) feature perturbation metrics outlined in our work. Note that the SHAP benchmark does not include any stability/robustness or fairness metrics. In contrast, OpenXAI benchmarks these metrics extensively across a variety of explanation methods (\eg LIME, Gradient-based methods).

\textbf{\emph{XAI-Bench}}~\cite{liu2021synthetic} constructed synthetic datasets with ground truth explanations to evaluate the faithfulness of a few explanation methods (\eg LIME, SHAP, MAPLE). However, recent research argued that their evaluation is unreliable, and predictive models learned using their synthetic datasets may not adhere to the ground truth explanations~\citep{faber2021comparing}. In addition, the aforementioned evaluation is rather limited in scope as synthetic datasets may not even be representative of real-world data~\citep{faber2021comparing}. In contrast, our work not only proposes a novel synthetic data generator that addresses the shortcomings of the synthetic datasets constructed in XAI-Bench but also facilitates the evaluation and benchmarking of the faithfulness, stability, as well as the fairness of 6 state-of-the-art explanation methods on 7 real-world datasets with no ground truth explanations.

In summary, our work makes the following key contributions:
\begin{itemize}
    \item We provide implementations and easy-to-use API interfaces for 11 metrics to evaluate the faithfulness, stability, and fairness of explanation methods, 7 of which have not been implemented in prior libraries or benchmarks -- the faithfulness metrics Feature Agreement (FA), Rank Agreement (RA), Sign Agreement (SA), Signed Rank Agreement (SRA), Pairwise Rank Agreement (PRA), and the stability metrics Relative Representation Stability (RRS) and Relative Output Stability (ROS).
    \item We also introduce a novel and flexible synthetic data generator to synthesize datasets of varying sizes, complexity, and dimensionality to facilitate the construction of reliable ground truth explanations in order to evaluate state-of-the-art explanation methods. Our synthetic data generator addresses the shortcomings of the prior synthetic benchmark (XAI-Bench) by generating synthetic datasets which encapsulate certain key properties, namely, unambiguously defined local neighborhoods, a clear description of feature importances in each local neighborhood, and feature independence. These properties, in turn, allow us to theoretically guarantee that any accurate model trained on our synthetic datasets will adhere to the ground truth explanations of the underlying data.
    \item We perform rigorous empirical benchmarking of 6 state-of-the-art feature attribution methods using our OpenXAI framework to determine which methods are effective w.r.t. each of the 11 evaluation metrics across 8 real-world/synthetic datasets, and 16 different predictive models. Note that none of the previously proposed libraries or benchmarks carry out such exhaustive benchmarking efforts across such a wide variety of metrics, models, and datasets.
\end{itemize}

\section{Overview of \name Framework}
\label{sec:overview}
\name provides a comprehensive programmatic environment 
with synthetic and real-world datasets, data processing functions, explainers, and evaluation metrics to rigorously and efficiently benchmark explanation methods. Below, we discuss each of these components in detail. 

\looseness=-1
\xhdr{1) Datasets and Predictive Models} The current release of our \name framework includes a collection of eight different synthetic and real-world datasets. While synthetic datasets allow us to construct ground truth explanations which can then be used to evaluate explanations output by state-of-the-art methods, real-world datasets (where it is typically hard to construct ground truth explanations) help us benchmark these methods in a more realistic manner suitable for practical applications~\cite{liu2021synthetic}. We would like to note that \name includes datasets that are widely employed in XAI research to evaluate the efficacy of newly proposed methods and study the behavior of existing methods~\citep{balagopalan2022road,dai2022fairness,dasgupta2022framework,dominguez2022adversarial,karimi2020algorithmic,slack2021reliable,upadhyay2021towards}.


\emph{Synthetic Datasets}: While prior research~\cite{liu2021synthetic,kim2021sanity} proposed methods to generate synthetic datasets and corresponding ground truth explanations, they all suffer from a significant drawback as demonstrated by~\citet{faber2021comparing} -- there is no guarantee that the models trained on these datasets will adhere to the ground truth explanations of the underlying data. This, in turn, implies that evaluating post hoc explanations using the above ground truth explanations would be incorrect since post hoc explanations are supposed to reliably explain the behavior of the underlying model, and not that of the underlying data. To illustrate, let us consider the case where we use  aforementioned methods to construct a synthetic dataset with features $A, B,$ $C$, and $D$ such that the ground truth labels only depend on features $A$ and $B$ i.e., the ground truth explanation of the underlying data indicates that features $A$ and $B$ are most important. If we train a model on this data and if features $A$ and $B$ are correlated with $C$ and $D$ respectively, then the resulting model may base its predictions on $C$ and $D$ (and not $A$ and $B$) and still be very accurate. If a post hoc explanation of this model then (correctly) indicates that the most important features of the model are $C$ and $D$, this explanation may be deemed incorrect if we compare it against the ground truth explanation of the underlying data. This problem further exacerbates as we increase the complexity of the ground truth labeling function~\cite{faber2021comparing}. 

To address the aforementioned challenges, we develop a novel synthetic data generation mechanism, \emph{SynthGauss}, which encapsulates three key properties, namely, feature independence, unambiguously-defined local neighborhoods, and a clear description of feature influence in each local neighborhood. Intuitively, this approach generates $K$ well-separated clusters where points in each cluster $k \in \{1, 2, \cdots, K\}$ are sampled from a Gaussian distribution $\mathcal{N}(\mu_k, \Sigma_k)$ where $u_k \in R^d$ is the mean and $\Sigma_k \in R^{d \times d}$ is the covariance matrix. While this parameterization is general enough to support the construction of synthetic datasets of $K$ clusters with varying means and covariances, we set the means of all the clusters such that the intracluster distances are significantly smaller than the intercluster distances, and we set the covariance matrices of all the clusters to identity. This ensures that all the features are independent, and local neighborhoods (clusters) are unambiguously defined. 
 
 We then generate ground truth labels for instances by first randomly sampling feature mask vectors $m_k \in \{0,1\}^d$ (vectors comprising of 0s and 1s) for each cluster $k$. The vector $m_k$ determines which features influence the ground truth labeling process for instances in cluster $k$ (a value of 1 indicates that the corresponding feature is influential). 
 We then randomly sample feature weight vectors $w_k \in R^{d}$ which capture the relative importance of each of the features in the labeling process of instances in each cluster $k$. The ground truth labels of instances in each cluster $k$ are then computed as a function (e.g., sigmoid) of the feature values of individual instances, and the dot product of the corresponding cluster's feature mask vector and weight vector i.e., $ m_k \odot w_k$. Complete pseudocode and other details of this generation process are included in the Appendix. Note that $m_k$ corresponds to the ground truth explanation for all instances in cluster $k$. 
 Since our generation process is designed to encapsulate feature independence, unambiguous definitions of local neighborhoods, and clear descriptions of feature influences, any accurate model trained on the resulting dataset will adhere to the ground truth explanations of the underlying data (See Theorem 1 in Appendix). 

 
\emph{Real-world Datasets}: In the current release of \name, we include seven real-world datasets that are highly diverse in terms of several key properties. They comprise of data spanning multiple real-world domains (e.g., finance, lending, healthcare, and criminal justice), varying dataset sizes (e.g., small vs.~large-scale), dimensionalities (e.g., low vs.~high dimensional), class imbalance ratios, and feature types (e.g., continuous vs.~discrete). We focus on tabular data in this release as such data is commonly encountered in real-world applications where explainability is critical~\cite{verma2020counterfactual}, and has also been widely studied in XAI literature~\cite{liu2021synthetic}. Table~\ref{tab:dataset} provides a summary of the real-world datasets currently included in \name. See Section~\ref{app:datasets} in the Appendix for detailed descriptions of individual datasets. While these real-world datasets are primarily drawn from prior research and existing repositories, \name provides comprehensive data loading and pre-processing capabilities to make these datasets \emph{XAI-ready} (further details below). 

\begin{table}[ht]
    \centering\small
    \renewcommand{\arraystretch}{0.9}
    \setlength{\tabcolsep}{1.2pt}
    \caption{
        \textbf{Summary of currently available datasets in \name.} Here, “Feature Types” denotes whether features in the dataset are discrete or continuous, “Feature Information” describes what kind of information is captured in the dataset, and “Balanced” denotes whether the dataset is balanced w.r.t. the predictive label.
    }
    {\begin{tabular}{P{0.2\linewidth} P{0.1\linewidth} P{0.1\linewidth}P{0.2\linewidth}P{0.25\linewidth}P{0.1\linewidth}}
    \toprule
    {Dataset} & {Size} & {\# Features} & {Feature Types} & {Feature Information} & {Balanced} \\
    \toprule
    {Synthetic Data} & {5,000} & {20} & {continuous} & {synthetic} & {\cmark} \\
    {German~Credit~\cite{Dua:2019}} & {1,000} & {20*} & {discrete, continuous} & {demographic, personal, financial} & {\xmark} \\
    {HELOC~\cite{HELOC}} & {9,871} & {23} & {continuous} & {demographic, financial} & {\cmark} \\
    {Adult~Income~\cite{yeh2009comparisons}} & {45,222} & {13} & {discrete, continuous} & {demographic, personal, education/employment, financial} & {\xmark} \\
    {COMPAS~\cite{jordan2015effect}} & {6,172} & {7} & {discrete, continuous} & {demographic, personal, criminal} & {\xmark} \\
    {Give Me Some Credit~\cite{GiveMeCredit}} & {102,209} & {10} & {discrete, continuous} & {demographic, personal, financial} & {\xmark} \\
    {Pima-Indians Diabetes~\cite{smith1988using}} & {768} & {8} & {discrete, continuous} & {demographic, healthcare} & {\xmark} \\
    {Framingham Heart Study~\cite{Framingh21}} & {3,658} & {15} & {continuous} & {demographic, healthcare} & {\xmark} \\
    \bottomrule
    \end{tabular}}
    {\scriptsize *expands to 60 features after one hot encoding of discrete features}
    \label{tab:dataset}
\end{table}

\emph{Dataloaders}: \name provides a \texttt{ReturnLoaders} function that can be used to load the aforementioned collection of synthetic and real-world datasets as well as any other custom datasets, and ensures that they are \emph{XAI-ready}. More specifically, this class takes as input the name of an existing \name dataset and outputs a train set which can then be used to train a predictive model, and a test set which can be used to generate local explanations of the trained model. 
The code snippet below shows how to import the \texttt{ReturnLoaders} function and load an existing \name dataset. 
\begin{mdframed}[hidealllines=true,backgroundcolor=blue!10]
    \small
    \texttt{\textcolor{darkgreen}{from} \textcolor{blue}{openxai.dataloader} \textcolor{darkgreen}{import} ReturnLoaders\\
    trainloader, testloader = ReturnLoaders(data\_name=\textcolor{red}{`german'}, download=\textcolor{red}{True})\\
    inputs, labels = next(iter(testloader))
}\end{mdframed}
\emph{Pre-trained models}: We train two classes of predictive models (artificial neural networks and logistic regression models) and incorporate them into the \name framework so that they can be readily used for benchmarking explanation methods.
The code snippet below shows how to load \name's pre-trained models using our \texttt{LoadModel} function.
\begin{mdframed}[hidealllines=true,backgroundcolor=blue!10]
    \small
    \texttt{\textcolor{darkgreen}{from} \textcolor{blue}{openxai.model} \textcolor{darkgreen}{import} LoadModel\\
    model = LoadModel(data\_name=\textcolor{red}{`german'}, ml\_model=\textcolor{red}{`ann'})
}\end{mdframed}

\looseness=-1
\xhdr{2) Explainers} 
\name provides ready-to-use implementations of six state-of-the-art feature attribution methods, namely, \LIME, \SHAP, \Grads, \GradtimesInput, \SmoothGrad, and \IntGrad. An implementation of a random baseline which randomly assigns importance values to each of the features, and returns these random assignments as explanations is also included. Our implementations of these methods build on other open-source libraries (e.g., Captum~\cite{kokhlikyan2020captum}) as well as their original implementations. 
While methods such as \LIME and \SHAP leverage perturbations of data instances and their corresponding model predictions to \emph{learn} a local explanation model,  they do not require access to the internals of the models or their gradients. On the other hand, \Grads, \GradtimesInput, \SmoothGrad, and \IntGrad require access to the gradients of the underlying models but do not need to repeatedly query the models for their predictions (see Table~\ref{tab:explanation} in Appendix for a brief summary of these methods). These differences influence the efficiency with which explanations can be generated by these methods. \name provides an abstract \texttt{Explainer} class which enables us to load existing explanation methods as well as integrate new explanation methods.
\begin{mdframed}[hidealllines=true,backgroundcolor=blue!10]
    \small
    \texttt{\textcolor{darkgreen}{from} \textcolor{blue}{openxai.explainer} \textcolor{darkgreen}{import} Explainer\\
    exp\_method = Explainer(method=\textcolor{red}{`shap'}, model=model)\\
    explanations = exp\_method.get\_explanations(inputs)
}\end{mdframed}
All the explanation methods included in \name are readily accessible through the \texttt{Explainer} class, and users just have to specify the method name in order to invoke the appropriate method and generate explanations as shown in the above code snippet. 
Users can easily incorporate their own custom explanation methods into the \name framework by extending the \texttt{Explainer} class and including the code for their methods in the \texttt{get\_explanations} function (see template above) of this class. They can then submit a request to incorporate their custom methods into \name.

\hide{
    \begin{wrapfigure}{r}{0.33\textwidth}
        \centering
        \includegraphics[width=\linewidth]{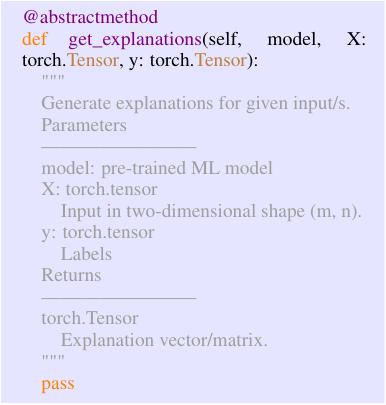}
        \caption{A template of \texttt{get\_} \texttt{explanations} function which enables users to incorporate custom explanation methods into \name.}
        \label{fig:exp-template}
        \end{wrapfigure}
}
\looseness=-1
\xhdr{3) Evaluation Metrics} 
\name provides implementations and ready-to-use APIs for a set of twenty-two quantitative metrics proposed by prior research to evaluate the faithfulness, stability, and fairness of explanation methods. \name is the first XAI benchmark to consider all the three aforementioned aspects of explanation reliability. More specifically, we include eight different metrics to measure explanation faithfulness (both with and without ground truth explanations)~\cite{krishna2022disagreement,petsiuk2018rise}, and three different metrics to measure stability~\cite{agarwal2022rethinking}. 
Below, we briefly describe these metrics. Detailed descriptions of all the metrics along with notation and equations are included in the Appendix.

a) \emph{Ground-truth Faithfulness}:
~\citet{krishna2022disagreement} recently proposed six evaluation metrics to capture the similarity between the top-K or a select set of features of any two feature attribution-based explanations. We leverage these metrics to capture the similarity between the explanations output by state-of-the-art methods and the ground-truth explanations constructed using our synthetic data generation process. These metrics and their definitions are given as follows: i) Feature Agreement (FA) which computes the fraction of top-K features that are common between a given post hoc explanation and the corresponding ground truth explanation, ii) Rank Agreement (RA) metric which measures the fraction of top-K features that are not only common between a given post hoc explanation and the corresponding ground truth explanation, but also have the same position in the respective rank orders, iii) Sign Agreement (SA) metric which computes the fraction of top-K features that are not only common between a given post hoc explanation and the corresponding ground truth explanation, but also share the same sign (direction of contribution) in both the explanations, iv) Signed Rank Agreement (SRA) metric which computes the fraction of top-K features that are not only common between a given post hoc explanation and the corresponding ground truth explanation, but also share the same feature attribution sign (direction of contribution) and position (rank) in both the explanations, v) Rank Correlation (RC) metric which computes Spearman’s rank correlation coefficient to measure the agreement between feature rankings provided by a given post hoc explanation and the corresponding ground truth explanation, and vi)  Pairwise Rank Agreement (PRA) metric which captures if the relative ordering of every pair of features is the same for a given post hoc explanation as well as the corresponding ground-truth explanation.

b) \emph{Predictive Faithfulness}: We leverage the metrics outlined by~\cite{petsiuk2018rise,dai2022fairness} to measure the faithfulness of an explanation when no ground truth is available. This metric, referred to as Prediction Gap on Important feature perturbation (PGI), computes the difference in prediction probability that results from perturbing the features deemed as influential by a given post hoc explanation. Higher values on this metric imply greater explanation faithfulness. We also consider the converse of this metric, Prediction Gap on Unimportant feature perturbation (PGU), which perturbs the unimportant features and measures the change in prediction probability. 

c) \emph{Stability}: We consider the metrics introduced by~\citet{alvarez2018robustness,agarwal2022rethinking} to measure how robust a given explanation is to small input perturbations. More specifically, we leverage the metrics Relative Input Stability (RIS), Relative Representation Stability (RRS), and Relative Output Stability (ROS) which measure the maximum change in explanation relative to changes in the inputs, model parameters, and output prediction probabilities respectively.

d) \emph{Fairness}: Following the work by~\citet{dai2022fairness}, we measure the fairness of post hoc explanations by averaging all the aforementioned metric values across instances in the majority and minority subgroups, and comparing the two estimates. If there is a huge difference in the two estimates, then we consider this to be evidence for unfairness. 

\looseness=-1
 Invoking the aforementioned metrics to benchmark an explanation methods is quite simple and the code snippet below describes how to invoke the PGI metric (where \texttt{kwargs} represents a dictionary containing the given metric's parameters). 
 
\begin{mdframed}[hidealllines=true,backgroundcolor=blue!10]
    \small
    \texttt{\textcolor{darkgreen}{from} \textcolor{blue}{openxai.evaluator} \textcolor{darkgreen}{import} Evaluator\linebreak
    metric\_evaluator = Evaluator(model, metric=\textcolor{red}{`PGI'})\linebreak
    score, mean\_score = metric\_evaluator.evaluate(**kwargs)
}
\end{mdframed}
\emph{\textbf{Benchmarking}}: As can be seen from the code snippets in this section, \name allows end users to easily benchmark explanation methods using just a few lines of code. To summarize the benchmarking process, let us consider a scenario where we would like to benchmark a new explanation method using \name's pre-trained neural network model and the German Credit dataset. First, we use \name's \texttt{ReturnLoaders} function to load the German Credit dataset. Second, we load the neural network model (`ann') using our \texttt{LoadModel} function. Third, we extend the \texttt{Explainer} class and incorporate the code for the new explanation method in the \texttt{get\_explanations} function of this class. Finally, we evaluate the new explanation method using any metric from the \texttt{Evaluator} class. 

\section{Benchmarking Analysis}
\label{sec:expt}
Next, we describe how we benchmark state-of-the-art explanation methods using our \name framework and also discuss the key findings of this benchmarking analysis. Code to reproduce all the results is available at \url{https://github.com/AI4LIFE-GROUP/OpenXAI}.

\looseness=-1
\xhdr{Experimental Setup} We benchmark all of the six state-of-the-art feature attribution methods currently available in our \name framework along with the random baseline, using the \texttt{openxai.Evaluator} module (See Section~\ref{sec:overview}), and for the first 1000 test instances of each dataset. 
Details about the hyperparameters used in our experiments are discussed in Section ~\ref{app:parameter} in the Appendix. Our \name framework currently has two pre-trained models, a logistic regression model and a deep neural network model, for each dataset. The neural network models have two fully connected hidden layers with 100 nodes in each layer, and they use ReLU activation functions and an output softmax layer. See Appendix~\ref{app:model} for details on model architectures, training and performance.

\xhdr{Faithfulness}
\label{sec:faith}
We evaluate the ground-truth and predictive faithfulness of explanations generated by state-of-the-art methods using both synthetic and real-world datasets.

\emph{Ground-truth faithfulness}: For logistic regression models, we evaluate ground-truth faithfulness by calculating the similarity between the generated explanations and the ground-truth explanations using the metrics discussed in Section~\ref{sec:overview}.
Results for various ground-truth faithfulness metrics are shown in Tables~\ref{tab:heloc_faith_lr}-\ref{tab:adult_faith_lr}, \ref{tab:pima_faith_lr}-\ref{tab:heart_faith_lr}, \ref{tab:gaussian_faith_lr}-\ref{tab:gmsc_faith_lr}. \Grads, \IntGrad, and \SmoothGrad produce explanations that achieve perfect scores on four ground-truth faithfulness metrics, viz. pairwise rank agreement (PRA), rank correlation (RC), feature agreement (FA), rank agreement (RA), sign agreement (SA), signed rank agreement (SRA) metrics, for all datasets. LIME outperforms other methods, approaching the performance of the above gradient-based explanations in many cases.
While illustrative in nature, these findings show how OpenXAI can help identify the limitations of existing explanation methods, which in turn can inform the design of new methods.

\emph{Predictive faithfulness}:
%
Tables~\ref{tab:heloc_faith_lr}-\ref{tab:adult_faith_lr},  \ref{tab:pima_faith_lr}-\ref{tab:heart_faith_lr}, \ref{tab:gaussian_faith_lr}-\ref{tab:gmsc_faith_lr} show results for logistic regression models, and Tables~\ref{tab:pima_faith_ann}-\ref{tab:heart_faith_ann}, \ref{tab:gaussian_faith_ann}-\ref{tab:gmsc_faith_ann} show results for neural network models on the PGI and PGU metrics implemented in \name (see Section~\ref{sec:overview} and Appendix~\ref{app:metric}). Overall, we find that \Grads, \SmoothGrad, and LIME explanations are most faithful to the underlying model and, on average across all datasets and models, outperform other feature-attribution methods on the PGI metric.
The PGU metric showed similar trends in terms of best performance, with \IntGrad on par with the leading metrics.

\begin{table}[t]
    \centering\small
    \renewcommand{\arraystretch}{0.9}
    \setlength{\tabcolsep}{1.5pt}
    \caption{
        \textbf{Ground-truth and predictive faithfulness results on the HELOC dataset for all explanation methods with LR model.} Shown are average and standard error metric values computed across 1000 test instances. $\uparrow$ indicates that higher values are better, and $\downarrow$ indicates that lower values are better. Values corresponding to best performance are bolded.
    }
    {\begin{tabular}{lcccccccc}
    \toprule
    {{Method}} & {PRA~($\uparrow$)} & {RC~($\uparrow$)} & {FA~($\uparrow$)} & {RA~($\uparrow$)} & {SA~($\uparrow$)} & {SRA~($\uparrow$)} & {PGU~($\downarrow$)} & {PGI~($\uparrow$)} \\
    \toprule
    \begin{tabular}[l]{@{}l@{}}{Random}\\{VanillaGrad}\\{IntegratedGrad}\\{\GradtimesInput}\\{\SmoothGrad}\\{\SHAP}\\{\LIME}
    \end{tabular} &
	\begin{tabular}[c]{@{}c@{}}
	{0.500}\std{0.00}\\
	\textbf{1.000}\std{0.00}\\
	\textbf{1.000}\std{0.00}\\
	{0.599}\std{0.00}\\
	\textbf{1.000}\std{0.00}\\
	{0.603}\std{0.00}\\
	{0.923}\std{0.00}\\
	\end{tabular} &
	\begin{tabular}[c]{@{}c@{}}
	{0.002}\std{0.01}\\
	\textbf{1.000}\std{0.00}\\
	\textbf{1.000}\std{0.00}\\
	{0.291}\std{0.01}\\
	\textbf{1.000}\std{0.00}\\
	{0.291}\std{0.01}\\
	{0.946}\std{0.00}\\
	\end{tabular} &
	\begin{tabular}[c]{@{}c@{}}
	{0.149}\std{0.00}\\
	\textbf{1.000}\std{0.00}\\
	\textbf{1.000}\std{0.00}\\
	{0.205}\std{0.00}\\
	\textbf{1.000}\std{0.00}\\
	{0.195}\std{0.00}\\
	{0.910}\std{0.00}\\
	\end{tabular} &
	\begin{tabular}[c]{@{}c@{}}
	{0.046}\std{0.00}\\
	\textbf{1.000}\std{0.00}\\
	\textbf{1.000}\std{0.00}\\
	{0.029}\std{0.00}\\
	\textbf{1.000}\std{0.00}\\
	{0.029}\std{0.00}\\
	{0.682}\std{0.01}\\
	\end{tabular} &
	\begin{tabular}[c]{@{}c@{}}
	{0.076}\std{0.00}\\
	\textbf{1.000}\std{0.00}\\
	\textbf{1.000}\std{0.00}\\
	{0.205}\std{0.00}\\
	\textbf{1.000}\std{0.00}\\
	{0.195}\std{0.00}\\
	{0.910}\std{0.00}\\
	\end{tabular} &
	\begin{tabular}[c]{@{}c@{}}
	{0.024}\std{0.00}\\
	\textbf{1.000}\std{0.00}\\
	\textbf{1.000}\std{0.00}\\
	{0.029}\std{0.00}\\
	\textbf{1.000}\std{0.00}\\
	{0.029}\std{0.00}\\
	{0.682}\std{0.01}\\
	\end{tabular} &
	\begin{tabular}[c]{@{}c@{}}
	{0.098}\std{0.00}\\
	\textbf{0.073}\std{0.00}\\
	\textbf{0.073}\std{0.00}\\
	{0.096}\std{0.00}\\
	\textbf{0.073}\std{0.00}\\
	{0.096}\std{0.00}\\
	\textbf{0.073}\std{0.00}\\
	\end{tabular} &
	\begin{tabular}[c]{@{}c@{}}
	{0.039}\std{0.00}\\
	\textbf{0.076}\std{0.00}\\
	\textbf{0.076}\std{0.00}\\
	{0.046}\std{0.00}\\
	\textbf{0.076}\std{0.00}\\
	{0.045}\std{0.00}\\
	\textbf{0.076}\std{0.00}\\
	\end{tabular}\\
    \bottomrule
    \end{tabular}}
    \label{tab:heloc_faith_lr}
\end{table}
\begin{table}[t]
    \centering\small
    \renewcommand{\arraystretch}{0.9}
    \setlength{\tabcolsep}{1.5pt}
    \caption{
        \textbf{Ground-truth and predictive faithfulness results on the Adult Income dataset for all explanation methods with LR model.} Shown are average and standard error metric values computed across 1000 test instances. $\uparrow$ indicates that higher values are better, and $\downarrow$ indicates that lower values are better. Values corresponding to best performance are bolded.
    }
    {\begin{tabular}{lcccccccc}
    \toprule
    {{Method}} & {PRA~($\uparrow$)} & {RC~($\uparrow$)} & {FA~($\uparrow$)} & {RA~($\uparrow$)} & {SA~($\uparrow$)} & {SRA~($\uparrow$)} & {PGU~($\downarrow$)} & {PGI~($\uparrow$)} \\
    \toprule
    \begin{tabular}[l]{@{}l@{}}{Random}\\{VanillaGrad}\\{IntegratedGrad}\\{\GradtimesInput}\\{\SmoothGrad}\\{\SHAP}\\{\LIME}
    \end{tabular} &
	\begin{tabular}[c]{@{}c@{}}
	{0.501}\std{0.00}\\
	\textbf{1.000}\std{0.00}\\
	\textbf{1.000}\std{0.00}\\
	{0.538}\std{0.00}\\
	\textbf{1.000}\std{0.00}\\
	{0.627}\std{0.00}\\
	{0.881}\std{0.00}\\
	\end{tabular} &
	\begin{tabular}[c]{@{}c@{}}
	{0.003}\std{0.01}\\
	\textbf{1.000}\std{0.00}\\
	\textbf{1.000}\std{0.00}\\
	{0.215}\std{0.00}\\
	\textbf{1.000}\std{0.00}\\
	{0.319}\std{0.01}\\
	{0.876}\std{0.00}\\
	\end{tabular} &
	\begin{tabular}[c]{@{}c@{}}
	{0.188}\std{0.01}\\
	{0.999}\std{0.00}\\
	\textbf{1.000}\std{0.00}\\
	{0.359}\std{0.00}\\
	{0.999}\std{0.00}\\
	{0.363}\std{0.00}\\
	{0.924}\std{0.00}\\
	\end{tabular} &
	\begin{tabular}[c]{@{}c@{}}
	{0.073}\std{0.00}\\
	{0.998}\std{0.00}\\
	\textbf{1.000}\std{0.00}\\
	{0.138}\std{0.01}\\
	{0.999}\std{0.00}\\
	{0.041}\std{0.00}\\
	{0.875}\std{0.00}\\
	\end{tabular} &
	\begin{tabular}[c]{@{}c@{}}
	{0.093}\std{0.00}\\
	{0.999}\std{0.00}\\
	\textbf{1.000}\std{0.00}\\
	{0.359}\std{0.00}\\
	{0.999}\std{0.00}\\
	{0.362}\std{0.00}\\
	{0.924}\std{0.00}\\
	\end{tabular} &
	\begin{tabular}[c]{@{}c@{}}
	{0.034}\std{0.00}\\
	{0.998}\std{0.00}\\
	\textbf{1.000}\std{0.00}\\
	{0.138}\std{0.01}\\
	{0.999}\std{0.00}\\
	{0.041}\std{0.00}\\
	{0.875}\std{0.00}\\
	\end{tabular} &
	\begin{tabular}[c]{@{}c@{}}
	{0.130}\std{0.00}\\
	{0.065}\std{0.00}\\
	{0.065}\std{0.00}\\
	{0.124}\std{0.00}\\
	{0.065}\std{0.00}\\
	{0.125}\std{0.00}\\
	\textbf{0.063}\std{0.00}\\
	\end{tabular} &
	\begin{tabular}[c]{@{}c@{}}
	{0.043}\std{0.00}\\
	{0.124}\std{0.00}\\
	{0.124}\std{0.00}\\
	{0.062}\std{0.00}\\
	{0.124}\std{0.00}\\
	{0.062}\std{0.00}\\
	\textbf{0.126}\std{0.00}\\
	\end{tabular}\\
    \bottomrule
    \end{tabular}}
    \label{tab:adult_faith_lr}
\end{table}

\xhdr{Stability}
\label{sec:stability}
Next, we examine the stability of explanation methods when the underlying models are logistic regression models in Tables ~\ref{tab:gaussian_stab_lr}-\ref{tab:german_stab_lr} below (additionally, Tables~\ref{tab:pima_stab_lr}-\ref{tab:heart_stab_lr}, \ref{tab:heloc_stab_lr}-~\ref{tab:gmsc_stab_lr} in the Appendix), and neural network models (Tables~\ref{tab:pima_stab_ann}-\ref{tab:heart_stab_ann}, 
\ref{tab:gaussian_stab_ann}-~\ref{tab:gmsc_stab_ann}). For logistic regression models, we consider only the RIS and ROS metrics for stability, since RRS is equivalent to ROS within this model class.
Overall, we see similar trends across explanation methods for each of the relative stability metrics at our chosen set of hyperparameters (see Appendix~\ref{app:parameter}).
First, \IntGrad tends to outperform all other feature-attribution methods with log average scores of 1.95 and 1.86 for RIS and ROS, respectively, when averaged across all datasets. \Grads and \GradtimesInput remain on roughly the same order of magnitude with log average scores of 3.16 and 3.37 for RIS, and 2.91 and 4.49 for ROS. The remaining methods, LIME, SHAP, \SmoothGrad and Random, score log RIS and ROS scores between roughly 10 and 15.

\vspace{10pt}
\begin{minipage}[c]{0.49\textwidth}
	\centering\small
    \renewcommand{\arraystretch}{0.9}
    \setlength{\tabcolsep}{2.0pt}
    \captionof{table}{
        \textbf{Stability results on the Synthetic dataset for all explanation methods with LR model.} Shown are log average and standard error metric values computed across 1000 test instances. $\uparrow$ indicates that higher values are better, and $\downarrow$ indicates that lower values are better. Values corresponding to best performance are bolded.
    }
    {\begin{tabular}{lcc}
    \toprule
    {{Method}} & {RIS~($\downarrow$)} & {ROS~($\downarrow$)} \\
    \toprule
    \begin{tabular}[l]{@{}l@{}}{Random}\\{VanillaGrad}\\{IntegratedGrad}\\{\GradtimesInput}\\{\SmoothGrad}\\{\SHAP}\\{\LIME}
    \end{tabular} &
	\begin{tabular}[c]{@{}c@{}}
	{11.58}\std{0.00}\\
	{2.49}\std{0.01}\\
	\textbf{1.17}\std{0.01}\\
	{2.50}\std{0.01}\\
	{9.69}\std{0.01}\\
	{9.35}\std{0.01}\\
	{9.10}\std{0.02}\\
	\end{tabular} &
	\begin{tabular}[c]{@{}c@{}}
	{12.34}\std{0.02}\\
	{0.85}\std{0.04}\\
	\textbf{-0.70}\std{0.02}\\
	{2.82}\std{0.04}\\
	{11.25}\std{0.02}\\
	{11.69}\std{0.02}\\
	{12.17}\std{0.04}\\
	\end{tabular}\\
    \bottomrule
    \end{tabular}}
    \label{tab:gaussian_stab_lr}
\end{minipage}
\quad
\begin{minipage}[c]{0.49\textwidth}
	\centering\small
    \renewcommand{\arraystretch}{0.9}
    \setlength{\tabcolsep}{2.0pt}
    \captionof{table}{
        \textbf{Stability results on the German Credit dataset for all explanation methods with LR model.} Shown are log average and standard error metric values computed across 1000 test instances. $\uparrow$ indicates that higher values are better, and $\downarrow$ indicates that lower values are better. Values corresponding to best performance are bolded.
    }
    {\begin{tabular}{lcc}
    \toprule
    {{Method}} & {RIS~($\downarrow$)} & {ROS~($\downarrow$)} \\
    \toprule
    \begin{tabular}[l]{@{}l@{}}{Random}\\{VanillaGrad}\\{IntegratedGrad}\\{\GradtimesInput}\\{\SmoothGrad}\\{\SHAP}\\{\LIME}
    \end{tabular} &
	\begin{tabular}[c]{@{}c@{}}
	{12.84}\std{0.00}\\
	{-0.01}\std{0.04}\\
	\textbf{-0.78}\std{0.03}\\
	{0.47}\std{0.01}\\
	{8.87}\std{0.03}\\
	{10.01}\std{0.07}\\
	{9.59}\std{0.01}\\
	\end{tabular} &
	\begin{tabular}[c]{@{}c@{}}
	{14.78}\std{0.04}\\
	{-0.21}\std{0.07}\\
	\textbf{-1.20}\std{0.05}\\
	{3.10}\std{0.09}\\
	{10.71}\std{0.07}\\
	{13.88}\std{0.10}\\
	{11.34}\std{0.05}\\
	\end{tabular}\\
    \bottomrule
    \end{tabular}}
    \label{tab:german_stab_lr}
\end{minipage}
\vspace{30pt}

\xhdr{Fairness}
\hide{Fairness results are shown in Figures~\ref{fig:german_disparity_lr}-\ref{fig:adult_disparity_lr}. Note that the disparity can be calculated on any faithfulness or stability metrics. Without loss of generality, we explore the disparity in predictive faithfulness across different subgroups in real-world datasets. First, the disparity analysis on prediction gap on unimportant features (PGU) in Figure~\ref{fig:german_disparity_lr} shows that the quality of explanations from \Grads and \IntGrad methods are inconsistent across different population subgroups (here, male vs. female). Second, similar results for Adult Income dataset in Figure~\ref{fig:adult_disparity_lr} observe a distinct gap in the predictive faithfulness performance across two sub-groups for all feature-attribution methods and \GradtimesInput generates the most consistent explanations across both subgroups (male vs. female). These results allude to a trade-off among different metrics where \GradtimesInput underperforms for faithfulness and stability metrics but outperforms for fairness metrics. See Appendix~\ref{app:results_ann} for similar results using ANN models.
}
To measure fairness of explanation methods, we compute the average metric values (for each of the aforementioned faithfulness and stability metrics) for different subgroups (e.g., male and female) in the dataset and compare them. Larger gaps between the metric values for different subgroups indicates higher disparities (unfairness). 
We present results for the PGU (see Section~\ref{sec:overview} and Appendix~\ref{app:metric}) metric.
The fairness analysis in Figures ~\ref{fig:german_disparity_lr} and \ref{fig:adult_disparity_lr} shows that there are disparities in the faithfulness of explanations (see Section~\ref{sec:overview}) output by several methods (most notably, \GradtimesInput and SHAP on the Adult Income dataset).
Results with NN models 
are included in Appendix~\ref{app:results_ann}.

\begin{minipage}{\textwidth}
  \begin{minipage}[b]{0.49\textwidth}
      \centering
      \includegraphics[width=0.99\textwidth,center]{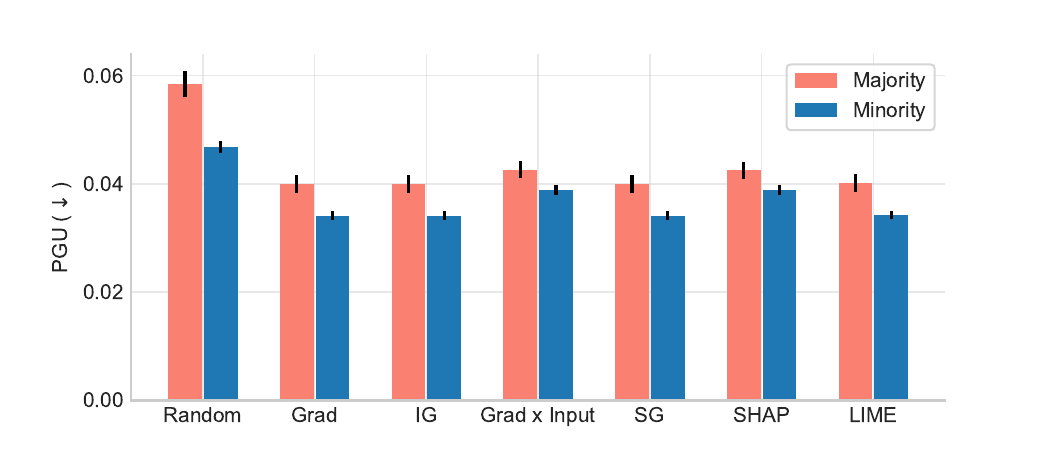}
      \captionof{figure}{\textbf{Fairness analysis of PGU metric on the German Credit dataset with LR model.} Shown are average and standard error values for majority (male) and minority (female) subgroups. Larger gaps between the values of majority and minority subgroups (\ie red and blue bars respectively) indicate higher disparities which are undesirable. 
      }
      \label{fig:german_disparity_lr}
  \end{minipage}
  \hfill
  \begin{minipage}[b]{0.49\textwidth}
      \centering
      \includegraphics[width=0.99\textwidth,center]{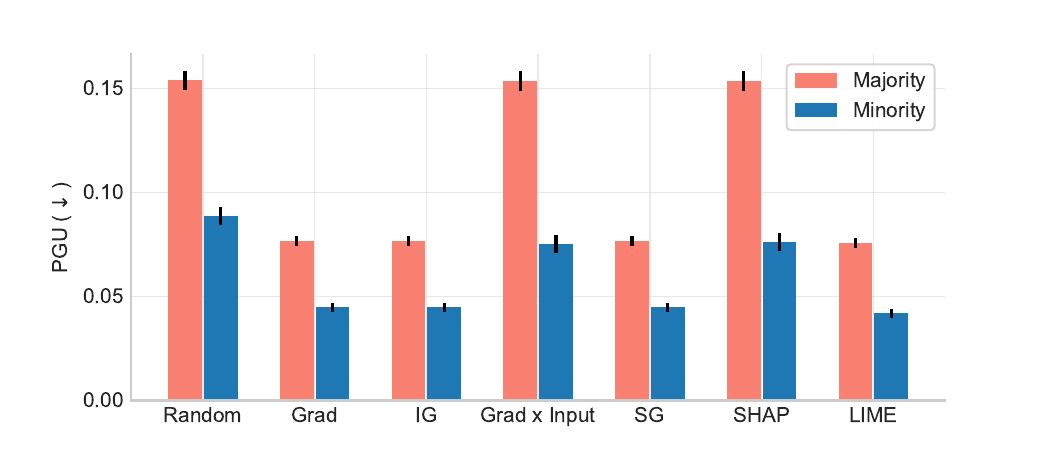}
      \captionof{figure}{\textbf{Fairness analysis of PGU metric on the Adult Income dataset with LR model.} Shown are average and standard error values for the majority (male) and minority (female) subgroups. Larger gaps between the values of majority and minority subgroups (\ie red and blue bars respectively) indicate higher disparities which are undesirable. }
      \label{fig:adult_disparity_lr}
  \end{minipage}
\end{minipage}

\section{Conclusion}
\label{sec:concl}
As post hoc explanations are increasingly being employed to aid decision makers and relevant stakeholders in various high-stakes applications, it becomes important to ensure that these explanations are reliable. To this end, we introduce \name, an open-source ecosystem comprising of XAI-ready datasets, implementations of state-of-the-art explanation methods, evaluation metrics, leaderboards and documentation to promote transparency and collaboration around evaluations of post hoc explanations. 
\name can readily be used to benchmark new explanation methods as well as incorporate them into our framework and leaderboards. By enabling systematic and efficient evaluation and benchmarking of existing and new explanation methods, OpenXAI can inform and accelerate new research in the emerging field of XAI. OpenXAI will be regularly updated with new datasets, explanation methods, and evaluation metrics, and welcomes input from the community.

\begin{ack}
    The authors would like to thank the anonymous reviewers for their helpful feedback and all the funding agencies listed below for supporting this work. This work is supported in part by the NSF awards \#IIS-2008461 and \#IIS-2040989, and research awards from Google, JP Morgan, Amazon, Harvard Data Science Initiative, and D$^3$ Institute at Harvard. HL would like to thank Sujatha and Mohan Lakkaraju for their continued support and encouragement. The views expressed here are those of the authors and do not reflect the official policy or position of the funding agencies.
\end{ack}











\bibliography{appendixbib,references}

\begin{thebibliography}{81}
\providecommand{\natexlab}[1]{#1}
\providecommand{\url}[1]{\texttt{#1}}
\expandafter\ifx\csname urlstyle\endcsname\relax
  \providecommand{\doi}[1]{doi: #1}\else
  \providecommand{\doi}{doi: \begingroup \urlstyle{rm}\Url}\fi

\bibitem[Fra()]{Framingh21}
Framingham heart study dataset | kaggle.
\newblock
  \url{https://www.kaggle.com/datasets/aasheesh200/framingham-heart-study-dataset}.
\newblock (Accessed on 08/15/2022).

\bibitem[sha()]{shapbench}
Shap benchmark.
\newblock URL \url{https://shap.readthedocs.io/en/latest/index.html}.

\bibitem[Agarwal and Nguyen(2020)]{agarwal2020explaining}
Chirag Agarwal and Anh Nguyen.
\newblock Explaining image classifiers by removing input features using
  generative models.
\newblock In \emph{ACCV}, 2020.

\bibitem[Agarwal et~al.(2022)Agarwal, Johnson, Pawelczyk, Krishna, Saxena,
  Zitnik, and Lakkaraju]{agarwal2022rethinking}
Chirag Agarwal, Nari Johnson, Martin Pawelczyk, Satyapriya Krishna, Eshika
  Saxena, Marinka Zitnik, and Himabindu Lakkaraju.
\newblock Rethinking stability for attribution-based explanations.
\newblock In \emph{ICLR 2022 Workshop on PAIR$^{2}$Struct}, 2022.

\bibitem[Agarwal et~al.(2021)Agarwal, Jabbari, Agarwal, Upadhyay, Wu, and
  Lakkaraju]{agarwal2021towards}
Sushant Agarwal, Shahin Jabbari, Chirag Agarwal, Sohini Upadhyay, Steven Wu,
  and Himabindu Lakkaraju.
\newblock Towards the unification and robustness of perturbation and gradient
  based explanations.
\newblock In \emph{ICML}, 2021.

\bibitem[Aivodji et~al.(2019)Aivodji, Arai, Fortineau, Gambs, Hara, and
  Tapp]{aivodji2019fairwashing}
Ulrich Aivodji, Hiromi Arai, Olivier Fortineau, S{\'e}bastien Gambs, Satoshi
  Hara, and Alain Tapp.
\newblock Fairwashing: the risk of rationalization.
\newblock In \emph{ICML}, 2019.

\bibitem[Alvarez-Melis and Jaakkola(2018)]{alvarez2018robustness}
David Alvarez-Melis and Tommi~S Jaakkola.
\newblock On the robustness of interpretability methods.
\newblock \emph{arXiv}, 2018.

\bibitem[Arrieta et~al.(2020)Arrieta, D{\'\i}az-Rodr{\'\i}guez, Del~Ser,
  Bennetot, Tabik, Barbado, Garc{\'\i}a, Gil-L{\'o}pez, Molina, Benjamins,
  et~al.]{arrieta2020explainable}
Alejandro~Barredo Arrieta, Natalia D{\'\i}az-Rodr{\'\i}guez, Javier Del~Ser,
  Adrien Bennetot, Siham Tabik, Alberto Barbado, Salvador Garc{\'\i}a, Sergio
  Gil-L{\'o}pez, Daniel Molina, Richard Benjamins, et~al.
\newblock Explainable artificial intelligence (xai): Concepts, taxonomies,
  opportunities and challenges toward responsible ai.
\newblock \emph{Information Fusion}, 2020.

\bibitem[Balagopalan et~al.(2022)Balagopalan, Zhang, Hamidieh, Hartvigsen,
  Rudzicz, and Ghassemi]{balagopalan2022road}
Aparna Balagopalan, Haoran Zhang, Kimia Hamidieh, Thomas Hartvigsen, Frank
  Rudzicz, and Marzyeh Ghassemi.
\newblock The road to explainability is paved with bias: Measuring the fairness
  of explanations.
\newblock \emph{arXiv}, 2022.

\bibitem[Bansal et~al.(2020)Bansal, Agarwal, and Nguyen]{bansal2020sam}
Naman Bansal, Chirag Agarwal, and Anh Nguyen.
\newblock Sam: The sensitivity of attribution methods to hyperparameters.
\newblock In \emph{CVPR}, 2020.

\bibitem[Barocas et~al.(2020)Barocas, Selbst, and Raghavan]{Barocas_2020}
Solon Barocas, Andrew Selbst, and Manish Raghavan.
\newblock The hidden assumptions behind counterfactual explanations and
  principal reasons.
\newblock In \emph{FAccT}, 2020.

\bibitem[Bastani et~al.(2017)Bastani, Kim, and
  Bastani]{bastani2017interpretability}
Osbert Bastani, Carolyn Kim, and Hamsa Bastani.
\newblock Interpretability via model extraction.
\newblock \emph{arXiv}, 2017.

\bibitem[Bien and Tibshirani(2009)]{bien2009classification}
Jacob Bien and Robert Tibshirani.
\newblock Classification by set cover: The prototype vector machine.
\newblock \emph{arXiv}, 2009.

\bibitem[Borisov et~al.(2021)Borisov, Leemann, Se{\ss}ler, Haug, Pawelczyk, and
  Kasneci]{borisov2021deep}
Vadim Borisov, Tobias Leemann, Kathrin Se{\ss}ler, Johannes Haug, Martin
  Pawelczyk, and Gjergji Kasneci.
\newblock Deep neural networks and tabular data: A survey.
\newblock \emph{arXiv}, 2021.

\bibitem[Caruana et~al.(2015)Caruana, Lou, Gehrke, Koch, Sturm, and
  Elhadad]{caruana15:intelligible}
Rich Caruana, Yin Lou, Johannes Gehrke, Paul Koch, Marc Sturm, and Noemie
  Elhadad.
\newblock Intelligible models for healthcare: Predicting pneumonia risk and
  hospital 30-day readmission.
\newblock In \emph{KDD}, 2015.

\bibitem[Chen et~al.(2022)Chen, Johnson, Topin, Plumb, and
  Talwalkar]{chen2022usecase}
Valerie Chen, Nari Johnson, Nicholay Topin, Gregory Plumb, and Ameet Talwalkar.
\newblock Use-case-grounded simulations for explanation evaluation.
\newblock \emph{arXiv}, 2022.

\bibitem[Covert et~al.(2021)Covert, Lundberg, and Lee]{covert2021explaining}
Ian Covert, Scott Lundberg, and Su-In Lee.
\newblock Explaining by removing: A unified framework for model explanation.
\newblock \emph{JMLR}, 2021.

\bibitem[Dai et~al.(2022)Dai, Upadhyay, Aivodji, Bach, and
  Lakkaraju]{dai2022fairness}
Jessica Dai, Sohini Upadhyay, Ulrich Aivodji, Stephen~H Bach, and Himabindu
  Lakkaraju.
\newblock Fairness via explanation quality: Evaluating disparities in the
  quality of post hoc explanations.
\newblock In \emph{AAAI Conference on AI, Ethics, and Society (AIES)}, 2022.

\bibitem[Dasgupta et~al.(2022)Dasgupta, Frost, and
  Moshkovitz]{dasgupta2022framework}
Sanjoy Dasgupta, Nave Frost, and Michal Moshkovitz.
\newblock Framework for evaluating faithfulness of local explanations.
\newblock \emph{arXiv}, 2022.

\bibitem[Dominguez-Olmedo et~al.(2022)Dominguez-Olmedo, Karimi, and
  Sch{\"o}lkopf]{dominguez2022adversarial}
Ricardo Dominguez-Olmedo, Amir~H Karimi, and Bernhard Sch{\"o}lkopf.
\newblock On the adversarial robustness of causal algorithmic recourse.
\newblock In \emph{ICML}. PMLR, 2022.

\bibitem[Doshi-Velez and Kim(2017)]{doshi2017towards}
Finale Doshi-Velez and Been Kim.
\newblock Towards a rigorous science of interpretable machine learning.
\newblock \emph{arXiv}, 2017.

\bibitem[Dua and Graff(2017)]{Dua:2019}
Dheeru Dua and Casey Graff.
\newblock {UCI} machine learning repository, 2017.
\newblock URL \url{http://archive.ics.uci.edu/ml}.

\bibitem[Elshawi et~al.(2019)Elshawi, Al-Mallah, and
  Sakr]{elshawi2019interpretability}
Radwa Elshawi, Mouaz~H Al-Mallah, and Sherif Sakr.
\newblock On the interpretability of machine learning-based model for
  predicting hypertension.
\newblock \emph{BMC medical informatics and decision making}, 2019.

\bibitem[Faber et~al.(2021)Faber, K.~Moghaddam, and
  Wattenhofer]{faber2021comparing}
Lukas Faber, Amin K.~Moghaddam, and Roger Wattenhofer.
\newblock When comparing to ground truth is wrong: On evaluating gnn
  explanation methods.
\newblock In \emph{KDD}, 2021.

\bibitem[FICO(2022)]{HELOC}
FICO.
\newblock Explainable machine learning challenge.
\newblock
  \url{https://community.fico.com/s/explainable-machine-learning-challenge?tabset-158d9=3},
  2022.
\newblock (Accessed on 05/23/2022).

\bibitem[Fokkema et~al.(2022)Fokkema, de~Heide, and van
  Erven]{fokkema2022attribution}
Hidde Fokkema, Rianne de~Heide, and Tim van Erven.
\newblock Attribution-based explanations that provide recourse cannot be
  robust.
\newblock \emph{arXiv}, 2022.

\bibitem[Freshcorn(2022)]{GiveMeCredit}
Bryce Freshcorn.
\newblock Give me some credit :: 2011 competition data | kaggle.
\newblock
  \url{https://www.kaggle.com/datasets/brycecf/give-me-some-credit-dataset},
  2022.
\newblock (Accessed on 05/23/2022).

\bibitem[Ghassemi et~al.(2021)Ghassemi, Oakden-Rayner, and
  Beam]{ghassemi2021false}
Marzyeh Ghassemi, Luke Oakden-Rayner, and Andrew~L Beam.
\newblock The false hope of current approaches to explainable artificial
  intelligence in health care.
\newblock \emph{The Lancet Digital Health}, 2021.

\bibitem[Ghorbani et~al.(2019)Ghorbani, Abid, and
  Zou]{ghorbani2019interpretation}
Amirata Ghorbani, Abubakar Abid, and James Zou.
\newblock Interpretation of neural networks is fragile.
\newblock In \emph{AAAI Conference on Artificial Intelligence}, 2019.

\bibitem[Guidotti et~al.(2018)Guidotti, Monreale, Ruggieri, Turini, Giannotti,
  and Pedreschi]{guidotti2018survey}
Riccardo Guidotti, Anna Monreale, Salvatore Ruggieri, Franco Turini, Fosca
  Giannotti, and Dino Pedreschi.
\newblock A survey of methods for explaining black box models.
\newblock \emph{ACM computing surveys (CSUR)}, 2018.

\bibitem[Han et~al.(2022)Han, Srinivas, and Lakkaraju]{han2022explanation}
Tessa Han, Suraj Srinivas, and Himabindu Lakkaraju.
\newblock Which explanation should i choose? a function approximation
  perspective to characterizing post hoc explanations.
\newblock \emph{arXiv}, 2022.

\bibitem[Hedstr{\"o}m et~al.(2022)Hedstr{\"o}m, Weber, Bareeva, Motzkus, Samek,
  Lapuschkin, and H{\"o}hne]{hedstrom2022quantus}
Anna Hedstr{\"o}m, Leander Weber, Dilyara Bareeva, Franz Motzkus, Wojciech
  Samek, Sebastian Lapuschkin, and Marina M-C H{\"o}hne.
\newblock Quantus: an explainable ai toolkit for responsible evaluation of
  neural network explanations.
\newblock \emph{arXiv}, 2022.

\bibitem[Hooker et~al.(2018)Hooker, Erhan, Kindermans, and
  Kim]{hooker2018evaluating}
Sara Hooker, Dumitru Erhan, Pieter-Jan Kindermans, and Been Kim.
\newblock Evaluating feature importance estimates.
\newblock \emph{arXiv}, 2018.

\bibitem[Ibrahim et~al.(2019)Ibrahim, Louie, Modarres, and
  Paisley]{ibrahim2019global}
Mark Ibrahim, Melissa Louie, Ceena Modarres, and John Paisley.
\newblock Global explanations of neural networks: Mapping the landscape of
  predictions.
\newblock \emph{CoRR, abs/1902.02384}, 2019.

\bibitem[Jesus et~al.(2021)Jesus, Bel{\'e}m, Balayan, Bento, Saleiro, Bizarro,
  and Gama]{jesus2021can}
S{\'e}rgio Jesus, Catarina Bel{\'e}m, Vladimir Balayan, Jo{\~a}o Bento, Pedro
  Saleiro, Pedro Bizarro, and Jo{\~a}o Gama.
\newblock How can i choose an explainer? an application-grounded evaluation of
  post-hoc explanations.
\newblock In \emph{FAccT}, 2021.

\bibitem[Jordan and Freiburger(2015)]{jordan2015effect}
Kareem~L Jordan and Tina~L Freiburger.
\newblock The effect of race/ethnicity on sentencing: Examining sentence type,
  jail length, and prison length.
\newblock In \emph{Journal of Ethnicity in Criminal Justice}. Taylor \&
  Francis, 2015.

\bibitem[Karimi et~al.(2019)Karimi, Barthe, Balle, and Valera]{MACE}
Amir-Hossein Karimi, Gilles Barthe, Borja Balle, and Isabel Valera.
\newblock Model-agnostic counterfactual explanations for consequential
  decisions.
\newblock \emph{arXiv}, 2019.

\bibitem[Karimi et~al.(2020{\natexlab{a}})Karimi, Sch{\"o}lkopf, and
  Valera]{karimi2020algorithmic}
Amir-Hossein Karimi, Bernhard Sch{\"o}lkopf, and Isabel Valera.
\newblock Algorithmic recourse: from counterfactual explanations to
  interventions.
\newblock \emph{CoRR, abs/2002.06278}, 2020{\natexlab{a}}.

\bibitem[Karimi et~al.(2020{\natexlab{b}})Karimi, von K{\"u}gelgen,
  Sch{\"o}lkopf, and Valera]{karimi2020causal}
Amir-Hossein Karimi, Julius von K{\"u}gelgen, Bernhard Sch{\"o}lkopf, and
  Isabel Valera.
\newblock Algorithmic recourse under imperfect causal knowledge: a
  probabilistic approach.
\newblock \emph{CoRR}, 2020{\natexlab{b}}.

\bibitem[Kaur et~al.(2020)Kaur, Nori, Jenkins, Caruana, Wallach, and
  Wortman~Vaughan]{kaur2020interpreting}
Harmanpreet Kaur, Harsha Nori, Samuel Jenkins, Rich Caruana, Hanna Wallach, and
  Jennifer Wortman~Vaughan.
\newblock Interpreting interpretability: Understanding data scientists' use of
  interpretability tools for machine learning.
\newblock In \emph{CHI Conference on Human Factors in Computing Systems}, 2020.

\bibitem[Kim et~al.(2021)Kim, Plumb, and Talwalkar]{kim2021sanity}
Joon~Sik Kim, Gregory Plumb, and Ameet Talwalkar.
\newblock Sanity simulations for saliency methods.
\newblock \emph{arXiv}, 2021.

\bibitem[Kokhlikyan et~al.(2020)Kokhlikyan, Miglani, Martin, Wang, Alsallakh,
  Reynolds, Melnikov, Kliushkina, Araya, Yan, and
  Reblitz-Richardson]{kokhlikyan2020captum}
Narine Kokhlikyan, Vivek Miglani, Miguel Martin, Edward Wang, Bilal Alsallakh,
  Jonathan Reynolds, Alexander Melnikov, Natalia Kliushkina, Carlos Araya, Siqi
  Yan, and Orion Reblitz-Richardson.
\newblock Captum: A unified and generic model interpretability library for
  pytorch, 2020.

\bibitem[Krishna et~al.(2022)Krishna, Han, Gu, Pombra, Jabbari, Wu, and
  Lakkaraju]{krishna2022disagreement}
Satyapriya Krishna, Tessa Han, Alex Gu, Javin Pombra, Shahin Jabbari, Steven
  Wu, and Himabindu Lakkaraju.
\newblock The disagreement problem in explainable machine learning: A
  practitioner's perspective.
\newblock \emph{arXiv}, 2022.

\bibitem[Lage et~al.(2019)Lage, Chen, He, Narayanan, Kim, Gershman, and
  Doshi-Velez]{lage2019evaluation}
Isaac Lage, Emily Chen, Jeffrey He, Menaka Narayanan, Been Kim, Sam Gershman,
  and Finale Doshi-Velez.
\newblock An evaluation of the human-interpretability of explanation.
\newblock \emph{arXiv}, 2019.

\bibitem[Lakkaraju and Bastani(2020)]{lakkaraju2020fool}
Himabindu Lakkaraju and Osbert Bastani.
\newblock ``how do i fool you?'' manipulating user trust via misleading black
  box explanations.
\newblock In \emph{AAAI Conference on AIES}, 2020.

\bibitem[Lakkaraju et~al.(2016)Lakkaraju, Bach, and
  Leskovec]{lakkaraju2016interpretable}
Himabindu Lakkaraju, Stephen~H Bach, and Jure Leskovec.
\newblock Interpretable decision sets: A joint framework for description and
  prediction.
\newblock In \emph{Proceedings of the 22nd ACM SIGKDD international conference
  on knowledge discovery and data mining}, pages 1675--1684, 2016.

\bibitem[Lakkaraju et~al.(2019)Lakkaraju, Kamar, Caruana, and
  Leskovec]{lakkaraju2019faithful}
Himabindu Lakkaraju, Ece Kamar, Rich Caruana, and Jure Leskovec.
\newblock Faithful and customizable explanations of black box models.
\newblock In \emph{Proceedings of the 2019 AAAI/ACM Conference on AI, Ethics,
  and Society}, pages 131--138, 2019.

\bibitem[Letham et~al.(2015)Letham, Rudin, McCormick, and
  Madigan]{letham2015interpretable}
Benjamin Letham, Cynthia Rudin, Tyler~H McCormick, and David Madigan.
\newblock Interpretable classifiers using rules and bayesian analysis: Building
  a better stroke prediction model.
\newblock \emph{The Annals of Applied Statistics}, 9\penalty0 (3):\penalty0
  1350--1371, 2015.

\bibitem[Linardatos et~al.(2021)Linardatos, Papastefanopoulos, and
  Kotsiantis]{linardatos2021explainable}
Pantelis Linardatos, Vasilis Papastefanopoulos, and Sotiris Kotsiantis.
\newblock Explainable ai: A review of machine learning interpretability
  methods.
\newblock \emph{Entropy}, 23\penalty0 (1):\penalty0 18, 2021.

\bibitem[Lipton(2016)]{lipton2016mythos}
Zachary~C Lipton.
\newblock The mythos of model interpretability.
\newblock \emph{CoRR, abs/1606.03490}, 2016.

\bibitem[Liu et~al.(2021)Liu, Khandagale, White, and
  Neiswanger]{liu2021synthetic}
Yang Liu, Sujay Khandagale, Colin White, and Willie Neiswanger.
\newblock Synthetic benchmarks for scientific research in explainable machine
  learning.
\newblock In \emph{NeurIPS Datasets and Benchmarks Track}, 2021.

\bibitem[Looveren and Klaise(2019)]{looveren2019interpretable}
Arnaud Looveren and Janis Klaise.
\newblock Interpretable counterfactual explanations guided by prototypes.
\newblock \emph{CoRR, abs/ 1907.02584}, 2019.

\bibitem[Lou et~al.(2012)Lou, Caruana, and Gehrke]{lou2012intelligible}
Yin Lou, Rich Caruana, and Johannes Gehrke.
\newblock Intelligible models for classification and regression.
\newblock In \emph{KDD}, 2012.

\bibitem[Lundberg and Lee(2017{\natexlab{a}})]{lundberg17:a-unified}
Scott~M Lundberg and Su-In Lee.
\newblock A unified approach to interpreting model predictions.
\newblock In I.~Guyon, U.~V. Luxburg, S.~Bengio, H.~Wallach, R.~Fergus,
  S.~Vishwanathan, and R.~Garnett, editors, \emph{Neural Information Processing
  Systems (NIPS)}, pages 4765--4774. Curran Associates, Inc.,
  2017{\natexlab{a}}.

\bibitem[Lundberg and Lee(2017{\natexlab{b}})]{lundberg2017unified}
Scott~M Lundberg and Su-In Lee.
\newblock A unified approach to interpreting model predictions.
\newblock In \emph{Advances in Neural Information Processing Systems}, pages
  4765--4774, 2017{\natexlab{b}}.

\bibitem[Murdoch et~al.(2019)Murdoch, Singh, Kumbier, Abbasi-Asl, and
  Yu]{murdoch2019definitions}
W~James Murdoch, Chandan Singh, Karl Kumbier, Reza Abbasi-Asl, and Bin Yu.
\newblock Definitions, methods, and applications in interpretable machine
  learning.
\newblock \emph{Proceedings of the National Academy of Sciences}, 2019.

\bibitem[Pawelczyk et~al.(2020)Pawelczyk, Broelemann, and
  Kasneci]{pawelczyk2020learning}
Martin Pawelczyk, Klaus Broelemann, and Gjergji Kasneci.
\newblock Learning model-agnostic counterfactual explanations for tabular data.
\newblock In \emph{WWW}, 2020.

\bibitem[Pawelczyk et~al.(2021)Pawelczyk, Bielawski, Van~den Heuvel, Richter,
  and Kasneci]{pawelczyk2021carla}
Martin Pawelczyk, Sascha Bielawski, Johan Van~den Heuvel, Tobias Richter, and
  Gjergji Kasneci.
\newblock Carla: A python library to benchmark algorithmic recourse and
  counterfactual explanation algorithms.
\newblock In \emph{NeurIPS Benchmark and Datasets Track}, 2021.

\bibitem[Petsiuk et~al.(2018)Petsiuk, Das, and Saenko]{petsiuk2018rise}
Vitali Petsiuk, Abir Das, and Kate Saenko.
\newblock Rise: Randomized input sampling for explanation of black-box models.
\newblock \emph{arXiv}, 2018.

\bibitem[Poursabzi-Sangdeh et~al.(2018)Poursabzi-Sangdeh, Goldstein, Hofman,
  Vaughan, and Wallach]{poursabzi2018manipulating}
Forough Poursabzi-Sangdeh, Daniel~G Goldstein, Jake~M Hofman, Jennifer~Wortman
  Vaughan, and Hanna Wallach.
\newblock Manipulating and measuring model interpretability.
\newblock \emph{CoRR}, 2018.

\bibitem[Poyiadzi et~al.(2020)Poyiadzi, Sokol, Santos-Rodriguez, De~Bie, and
  Flach]{FACE}
Rafael Poyiadzi, Kacper Sokol, Raul Santos-Rodriguez, Tijl De~Bie, and Peter
  Flach.
\newblock {FACE}: Feasible and actionable counterfactual explanations.
\newblock In \emph{AAAI Conference on AIES}, 2020.

\bibitem[Ribeiro et~al.(2016)Ribeiro, Singh, and Guestrin]{ribeiro16:kdd}
Marco~Tulio Ribeiro, Sameer Singh, and Carlos Guestrin.
\newblock "why should i trust you?": Explaining the predictions of any
  classifier.
\newblock In \emph{KDD}, 2016.

\bibitem[Ribeiro et~al.(2018)Ribeiro, Singh, and Guestrin]{ribeiro2018anchors}
Marco~Tulio Ribeiro, Sameer Singh, and Carlos Guestrin.
\newblock Anchors: High-precision model-agnostic explanations.
\newblock In \emph{AAAI}, 2018.

\bibitem[Rudin(2019)]{rudin2019stop}
Cynthia Rudin.
\newblock Stop explaining black box machine learning models for high stakes
  decisions and use interpretable models instead.
\newblock \emph{Nature Machine Intelligence}, 2019.

\bibitem[Selvaraju et~al.(2017)Selvaraju, Cogswell, Das, Vedantam, Parikh, and
  Batra]{selvaraju2017grad}
Ramprasaath~R Selvaraju, Michael Cogswell, Abhishek Das, Ramakrishna Vedantam,
  Devi Parikh, and Dhruv Batra.
\newblock Grad-cam: Visual explanations from deep networks via gradient-based
  localization.
\newblock In \emph{ICCV}, 2017.

\bibitem[Shrikumar et~al.(2017)Shrikumar, Greenside, and
  Kundaje]{shrikumar2017learning}
Avanti Shrikumar, Peyton Greenside, and Anshul Kundaje.
\newblock Learning important features through propagating activation
  differences.
\newblock In \emph{ICML}, 2017.

\bibitem[Simonyan et~al.(2014)Simonyan, Vedaldi, and
  Zisserman]{simonyan2013saliency}
Karen Simonyan, Andrea Vedaldi, and Andrew Zisserman.
\newblock Deep inside convolutional networks: Visualising image classification
  models and saliency maps.
\newblock In \emph{ICLR}, 2014.

\bibitem[Slack et~al.(2020)Slack, Hilgard, Jia, Singh, and
  Lakkaraju]{slack2020fooling}
Dylan Slack, Sophie Hilgard, Emily Jia, Sameer Singh, and Himabindu Lakkaraju.
\newblock Fooling lime and shap: Adversarial attacks on post hoc explanation
  methods.
\newblock In \emph{AAAI Conference on AIES}, 2020.

\bibitem[Slack et~al.(2021)Slack, Hilgard, Singh, and
  Lakkaraju]{slack2021reliable}
Dylan Slack, Anna Hilgard, Sameer Singh, and Himabindu Lakkaraju.
\newblock Reliable post hoc explanations: Modeling uncertainty in
  explainability.
\newblock \emph{NeurIPS}, 2021.

\bibitem[Smilkov et~al.(2017)Smilkov, Thorat, Kim, Vi{\'e}gas, and
  Wattenberg]{smilkov2017smoothgrad}
Daniel Smilkov, Nikhil Thorat, Been Kim, Fernanda Vi{\'e}gas, and Martin
  Wattenberg.
\newblock Smoothgrad: removing noise by adding noise.
\newblock \emph{arXiv}, 2017.

\bibitem[Smith et~al.(1988)Smith, Everhart, Dickson, Knowler, and
  Johannes]{smith1988using}
Jack~W Smith, James~E Everhart, WC~Dickson, William~C Knowler, and Robert~Scott
  Johannes.
\newblock Using the adap learning algorithm to forecast the onset of diabetes
  mellitus.
\newblock In \emph{Proceedings of the annual symposium on computer application
  in medical care}, page 261. American Medical Informatics Association, 1988.

\bibitem[Sundararajan et~al.(2017)Sundararajan, Taly, and
  Yan]{sundararajan2017axiomatic}
Mukund Sundararajan, Ankur Taly, and Qiqi Yan.
\newblock Axiomatic attribution for deep networks.
\newblock In \emph{ICML}, 2017.

\bibitem[Upadhyay et~al.(2021)Upadhyay, Joshi, and
  Lakkaraju]{upadhyay2021towards}
Sohini Upadhyay, Shalmali Joshi, and Himabindu Lakkaraju.
\newblock Towards robust and reliable algorithmic recourse.
\newblock \emph{NeurIPS}, 2021.

\bibitem[Ustun et~al.(2019)Ustun, Spangher, and Liu]{ustun2019actionable}
Berk Ustun, Alexander Spangher, and Yang Liu.
\newblock Actionable recourse in linear classification.
\newblock In \emph{FAccT}, 2019.

\bibitem[Verma et~al.(2020)Verma, Dickerson, and
  Hines]{verma2020counterfactual}
Sahil Verma, John Dickerson, and Keegan Hines.
\newblock Counterfactual explanations for machine learning: A review.
\newblock \emph{arXiv}, 2020.

\bibitem[Wachter et~al.(2017)Wachter, Mittelstadt, and
  Russell]{wachter2017counterfactual}
Sandra Wachter, Brent Mittelstadt, and Chris Russell.
\newblock Counterfactual explanations without opening the black box: Automated
  decisions and the {GDPR}.
\newblock \emph{Harvard Journal of Law \& Technology}, 31:\penalty0 841, 2017.

\bibitem[Wang and Rudin(2015)]{wang2015falling}
Fulton Wang and Cynthia Rudin.
\newblock Falling rule lists.
\newblock In \emph{Artificial Intelligence and Statistics}, pages 1013--1022.
  PMLR, 2015.

\bibitem[Whitmore et~al.(2016)Whitmore, George, and
  Hudson]{whitmore2016mapping}
Leanne~S Whitmore, Anthe George, and Corey~M Hudson.
\newblock Mapping chemical performance on molecular structures using locally
  interpretable explanations.
\newblock \emph{CoRR, abs/1611.07443}, 2016.

\bibitem[Yeh and Lien(2009)]{yeh2009comparisons}
I-Cheng Yeh and Che-hui Lien.
\newblock The comparisons of data mining techniques for the predictive accuracy
  of probability of default of credit card clients.
\newblock In \emph{Expert Systems with Applications}, 2009.

\bibitem[Zeng et~al.(2017)Zeng, Ustun, and Rudin]{zeng2017interpretable}
Jiaming Zeng, Berk Ustun, and Cynthia Rudin.
\newblock Interpretable classification models for recidivism prediction.
\newblock \emph{Journal of the Royal Statistical Society: Series A (Statistics
  in Society)}, 2017.

\bibitem[Zhou et~al.(2021)Zhou, Gandomi, Chen, and
  Holzinger]{zhou2021evaluating}
Jianlong Zhou, Amir~H Gandomi, Fang Chen, and Andreas Holzinger.
\newblock Evaluating the quality of machine learning explanations: A survey on
  methods and metrics.
\newblock \emph{Electronics}, 10\penalty0 (5):\penalty0 593, 2021.

\end{thebibliography}
\bibliographystyle{plainnat}

\newpage
\appendix
\section{Evaluation Metrics}
\label{app:metric}

\looseness=-1
Here, we describe different evaluation metrics that we included in the first iteration of \name. More specifically, we discuss various metrics (and their implementations) for evaluating the faithfulness, stability, and fairness of explanations generated using a given feature attribution method. 

\xhdr{1) Faithfulness} To measure how faithfully a given explanation mimics the underlying model, prior work has either leveraged synthetic datasets to obtain ground truth explanations or measured the differences in predictions when feature values are perturbed~\citep{petsiuk2018rise}. Here, we discuss the two broad categories of faithfulness metrics included in \name, namely ground-truth and predictive faithfulness.

\emph{a) Ground-truth Faithfulness.} 
\name leverages the following metrics outlined by~\citet{krishna2022disagreement} to calculate the agreement between ground-truth explanations (i.e., coefficients of logistic regression models) and explanations generated by state-of-the-art methods.
\begin{itemize}
    \item \emph{Feature Agreement (FA)} metric computes the fraction of top-$K$ features that are common between a given post hoc explanation and the corresponding ground truth explanation.
    \item \emph{Rank Agreement (RA)} metric measures the fraction of top-$K$ features that are not only common between a given post hoc explanation and the corresponding ground truth explanation, but also have the same position in the respective rank orders.
\item \emph{Sign Agreement (SA)} metric computes the fraction of top-$K$ features that are not only common
between a given post hoc explanation and the corresponding ground truth explanation, but also share the same sign (direction of contribution) in both the explanations. 
\item \emph{Signed Rank Agreement (SRA)} metric computes the fraction of top-$K$
features that are not only common between a given post hoc explanation and the corresponding ground truth explanation, but also share the
same feature attribution sign (direction of contribution) and position (rank) in both the explanations. 
\item \emph{Rank Correlation (RC)} metric computes the Spearman’s rank correlation
coefficient to measure the agreement between feature rankings provided by a given post hoc explanation and the corresponding ground truth explanation. 
\item \emph{Pairwise Rank Agreement (PRA)} metric captures if the relative ordering of every pair of features is the same for a given post hoc explanation as well as the corresponding ground truth explanation i.e., if feature A is more important than B according to one explanation, then the same should
be true for the other explanation. More specifically, this metric computes the fraction of feature pairs for
which the relative ordering is the same between the two explanations.
\end{itemize}

\looseness=-1
\xhdr{Reported metric values} While the aforementioned metrics quantify the ground-truth faithfulness of individual explanations, we report a single value corresponding to each explanation method and dataset to facilitate easy comparison of state-of-the-art methods. To this end, we adopt a similar strategy as that of~\citet{krishna2022disagreement} and compute the aforementioned metrics for each instance in the test data and then average these values. To arrive at the reported values of FA, RA, SA, and SRA metrics, instead of setting a specific value for $K$, we do the above computation for all possible values of $K$ and then plot the different values of $K$ (on the x-axis) and the corresponding averaged metric values (on the y-axis), and calculate the area under the resulting curve (AUC). On the other hand, in case of RC and PRA metrics, we set $K$ to the total number of features and compute the averaged metric values (across all test instances) as discussed above.

\emph{b) Predictive Faithfulness.} Following \citet{petsiuk2018rise}, \name includes two complementary predictive faithfulness metrics: i) Prediction Gap on Important feature perturbation (PGI) which measures the difference in prediction probability that results from perturbing the features deemed as influential by a given post hoc explanation, and ii) Prediction Gap on Unimportant feature perturbation (PGU) which measures the difference in prediction probability that results from perturbing the features deemed as unimportant by a given post hoc explanation. 

For a given instance $\mathbf{x}$, we first obtain the prediction probability $\hat{y}$ output by the underlying model $f$, \ie $\hat{y}=f(\mathbf{x})$. Let $e_{\mathbf{x}}$ be an explanation for the model prediction of $\mathbf{x}$. In case of PGU, we then generate a perturbed instance $\mathbf{x}'$ in the local neighborhood of $\mathbf{x}$ by holding the top-$k$ features constant, and slightly perturbing the values of all the other features by adding a small amount of Gaussian noise. In case of PGI, we generate a perturbed instance $\mathbf{x}'$ in the local neighborhood of $\mathbf{x}$ by slightly perturbing the values of the top-$k$ features by adding a small amount of Gaussian noise, and holding all the other features constant. 
Finally, we compute the expected value of the prediction difference between the original and perturbed instances as:

\begin{equation}
    \textup{PGI}(\mathbf{x}, f, e_{\mathbf{x}}, k) = \mathbb{E}_{\mathbf{x}' \sim \textnormal{perturb($\mathbf{x}$, $e_{\mathbf{x}}$, top-$K$)}}[|\hat{y} - f(\mathbf{x}')|],
    \label{eq:pred_preserve}
\end{equation}

\begin{equation}
    \textup{PGU}(\mathbf{x}, f, e_{\mathbf{x}}, k) = \mathbb{E}_{\mathbf{x}' \sim \textnormal{perturb($\mathbf{x}$, $e_{\mathbf{x}}$, non top-$K$)}}[|\hat{y} - f(\mathbf{x}')|],
    \label{eq:pred_preserve_inv}
\end{equation}

where perturb($\cdot$) returns the noisy versions of $\mathbf{x}$ as described above. 

\xhdr{Reported metric values} Similar to the ground-truth faithfulness metrics, we report PGI and PGU by calculating the AUC over all values of $K$.


\xhdr{2) Stability} 
We leverage the metrics Relative Input Stability (RIS), Relative Representation Stability (RRS), and Relative Output Stability (ROS)~\cite{agarwal2022rethinking} which measure the maximum change in explanation relative to changes in the inputs, internal representations learned by the model, and output prediction probabilities respectively. These metrics can be written formally as:
\hide{
\citet{alvarez2018robustness} formalizes the first stability metric for feature-attribution methods, arguing that explanations should be robust to local perturbations of an input. To evaluate the stability of an explanation, for instance, $\mathbf{x}$, perturbed instances $\mathbf{x}'$ are generated by adding infinitesimally small noise to the clean instance $\mathbf{x}$ such that $\hat{y}_{\mathbf{x}} = \hat{y}_{\mathbf{x}'}$:
}
\begin{equation}
    \textup{RIS} = \frac{||\mathbf{x}||_p}{||\mathbf{e}_{\mathbf{x}}||_p}\max_{\mathbf{x}'}\frac{||\mathbf{e}_{\mathbf{x}} - \mathbf{e}_{\mathbf{x}'}||_p}{||\mathbf{x} - \mathbf{x}'||_p},~\forall \mathbf{x}'~\textup{s.t.}~\mathbf{x}'\in\mathcal{N}_{\mathbf{x}};~\hat{y}_{\mathbf{x}}=\hat{y}_{\mathbf{x}'}
    \label{eq:old_stab1}
\end{equation}

where $\mathcal{N}_{\mathbf{x}}$ is a neighborhood of instances $\mathbf{x}'$ around $\mathbf{x}$, and $\mathbf{e}_{\mathbf{x}}$ and $\mathbf{e}_{\mathbf{x'}}$ denote the explanations corresponding to instances $\mathbf{x}$ and  $\mathbf{x}'$, respectively. The numerator of the metric measures the $l_p$ norm of the percent change of explanation $\mathbf{e}_{\mathbf{x}'}$  on the perturbed
instance $\mathbf{x}'$ with respect to the explanation $\mathbf{e}_\mathbf{x}$ on the original point $\mathbf{x}$. The denominator measures the $l_p$ norm between the
(normalized) inputs $\mathbf{x}$ and $\mathbf{x}'$,
and the max term in the denominator prevents division by zero. 
\begin{equation}
    \textup{RRS} = \frac{||\mathcal{L}_{\mathbf{x}}||_p}{||\mathbf{e}_{\mathbf{x}}||_p} \max_{\mathbf{x}'}\frac{ {||\mathbf{e}_{\mathbf{x}} - \mathbf{e}_{\mathbf{x}'}||_p}}{ ||\mathcal{L}_{\mathbf{x}}{-}\mathcal{L}_{\mathbf{x}'}||_p},~\forall \mathbf{x}'~~\textup{s.t.}~~\mathbf{x}'\in\mathcal{N}_{\mathbf{x}};~~\hat{y}_{\mathbf{x}}{=}\hat{y}_{\mathbf{x}'}
    \label{eq:rrs}
\end{equation}

where $\mathcal{L}(\cdot)$ denotes the internal representations learned by the model. 
Without loss of generality, we use a two layer neural network model in our experiments and use the output of the first layer for computing RRS. 
Similarly, we can also define ROS as:
\begin{equation}
    \textup{ROS} = \frac{||f({\mathbf{x}})||_p}{||\mathbf{e}_{\mathbf{x}}||_p}\max_{\mathbf{x}'}\frac{||\mathbf{e}_{\mathbf{x}} - \mathbf{e}_{\mathbf{x}'}||_p}{||f({\mathbf{x}}){-}f({\mathbf{x}'})||_p},~\forall \mathbf{x}'~~\textup{s.t.}~~\mathbf{x}'\in\mathcal{N}_{\mathbf{x}};~~\hat{y}_{\mathbf{x}}{=}\hat{y}_{\mathbf{x}'},
    \label{eq:ros}
\end{equation}
where $f(\mathbf{x})$ and $f(\mathbf{x}')$ are the output prediction probabilities for $\mathbf{x}$ and $\mathbf{x}'$, respectively. To the best of our knowledge, \name is the first benchmark to incorporate all the above stability metrics.

\xhdr{Reported metric values}. We compute the aforementioned metrics for each instance in the test set, and then average these values. 

\xhdr{3) Fairness} It is important to ensure that there are no significant disparities between the reliability of post hoc explanations corresponding to instances in the majority and the minority subgroups. To this end, we average all the aforementioned metric values across instances in the majority and minority subgroups, and then compare the two estimates to check if there are significant disparities~\cite{dai2022fairness}. See Figures~\ref{fig:german_disparity_lr} to~\ref{fig:adult_disparity_lr} for logistic regression models, and Figures~\ref{fig:german_disparity_ann} to~\ref{fig:adult_disparity_ann} for neural network models.

\section{Additional Related Work}
Our work builds on the vast literature on model interpretability and explainability. Below is an overview of additional works that were not included in Section~\ref{sec:intro} due to space constraints.

\xhdr{Inherently Interpretable Models and Post hoc Explanations} Many approaches learn inherently interpretable models such as rule lists \citep{zeng2017interpretable, wang2015falling}, decision trees and decision lists \citep{letham2015interpretable}, 
and others 
\citep{lakkaraju2016interpretable, bien2009classification, lou2012intelligible, caruana15:intelligible}.
However, complex models such as deep neural networks often achieve higher accuracy than simpler  models~\cite{ribeiro16:kdd}.
Thus, there has been significant interest in constructing post hoc explanations to understand their behavior. To this end, several techniques have been proposed in recent literature to construct \emph{post hoc explanations} of complex decision models.
For instance, LIME, SHAP, Anchors, BayesLIME, and BayesSHAP~\cite{ribeiro16:kdd,lundberg2017unified,ribeiro2018anchors,slack2021reliable} are considered \emph{perturbation-based local} explanation methods because they leverage perturbations of individual instances to construct interpretable local approximations (e.g., linear models). 
On the other hand, methods such as Vanilla Gradients, \GradtimesInput, SmoothGrad, Integrated Gradients, and GradCAM~\cite{simonyan2013saliency, sundararajan2017axiomatic, selvaraju2017grad,smilkov2017smoothgrad} are referred to as \emph{gradient-based local} explanation methods since they leverage gradients computed with respect to input features of individual instances to explain individual model predictions. 

There has also been recent work on constructing \emph{counterfactual explanations} which capture what changes need to be made to a given instance in order to flip its prediction~\cite{wachter2017counterfactual,ustun2019actionable,MACE,FACE,looveren2019interpretable,Barocas_2020,karimi2020algorithmic,pawelczyk2020learning,karimi2020causal,pawelczyk2021carla}. Such explanations can be leveraged to provide recourse to individuals negatively impacted by algorithmic decisions.
An alternate class of methods referred to as \emph{global} explanation methods attempt to summarize the behavior of black-box models as a whole rather than in relation to individual data points \citep{lakkaraju2019faithful,bastani2017interpretability}. A more detailed treatment of this topic is provided in other comprehensive survey articles~\citep{arrieta2020explainable, guidotti2018survey, murdoch2019definitions, linardatos2021explainable,covert2021explaining}. 

In this work, we focus primarily on \textit{local feature attribution-based post hoc explanation methods} i.e., explanation methods which attempt to explain individual model predictions by outputting a vector of feature importances. More specifically, the goal of this work is to enable systematic benchmarking of these methods in an efficient and transparent manner.

\xhdr{Evaluating Post hoc Explanations}
\hide{
Prior research has proposed several metrics to determine if an explanation is reliable. 
An extensive survey of metrics for evaluating explanation methods can be found in \citet{zhou2021evaluating}, who propose two high-level goals of explanation methods: interpretability (the clarity, simplicity, and broadness of the explanations), 
and fidelity (the completeness and soundness of explanations)~\cite{carvalho2019machine, gilpin2018explaining}. 
\citet{liu2021synthetic} provide a synthetic benchmark for explanation evaluation which includes implementations of several metrics, as well as a discussion of how to choose metrics for evaluation. Furthermore, several prior works proposed different metrics for evaluating various aspects of explanation quality, such as fidelity, stability, consistency, and sparsity~\cite{liu2021synthetic,petsiuk2018rise,slack2021reliable,zhou2021evaluating,ghorbani2019interpretation,alvarez2018robustness, hooker2018evaluating, lakkaraju2019faithful}. 
Recent research further leveraged the aforementioned properties and metrics to theoretically and empirically analyze the behavior of popular post hoc explanation methods~\cite{ghorbani2019interpretation, slack2019can,dombrowski2019explanations,adebayo2018sanity,alvarez2018robustness,levine2019certifiably,pmlr-v119-chalasani20a,agarwal2021towards}. }
In addition to the quantitative metrics designed to evaluate the \emph{reliability} of post hoc explanation methods~\cite{liu2021synthetic,zhou2021evaluating,alvarez2018robustness,dai2022fairness} (See Section~\ref{sec:intro}), prior works have also introduced 
\textit{human-grounded} approaches to evaluate the \emph{interpretability} of explanations generated by these methods~\citep{doshi2017towards}. 
For example, \citet{lakkaraju2020fool} carry out a user study to understand if misleading explanations can fool domain experts into deploying racially biased models, while \citet{kaur2020interpreting} find that explanations are often over-trusted and misused. 
Similarly, \citet{poursabzi2018manipulating} find that supposedly-interpretable models can lead to a decreased ability to detect and correct model mistakes, possibly due to information overload.
\citet{jesus2021can} introduce a method to compare explanation methods based on how subject matter experts perform on specific tasks with the help of explanations.
\citet{lage2019evaluation} use insights from rigorous human-subject experiments to inform regularizers used in explanation algorithms. In contrast to the aforementioned research, our work leverages twenty-two different state-of-the-art quantitative metrics to systematically benchmark the reliability (and not interpretability) of post hoc explanation methods.

\vspace{4pt}\noindent\textbf{Limitations and Vulnerabilities of Post hoc Explanations.}
Various quantitative metrics proposed in literature (See Section~\ref{sec:intro}) were also leveraged to analyze the behavior of post hoc explanation methods and their vulnerabilities---e.g.,~\citet{ghorbani2019interpretation} and~\citet{slack2020fooling} demonstrated that methods such as LIME and SHAP may result in explanations that are not only inconsistent and unstable, but also prone to adversarial attacks. 
Furthermore~\citet{lakkaraju2020fool} and~\citet{slack2020fooling} showed that explanations which do not accurately represent the importance of sensitive attributes (e.g., race, gender) could potentially mislead end users into believing that the underlying models are fair when they are not~\cite{lakkaraju2020fool,slack2020fooling,aivodji2019fairwashing}. This, in turn, could lead to the deployment of unfair models in critical real world applications.
There is also some discussion about whether models which are not inherently interpretable ought to be used in high-stakes decisions at all.
\citet{rudin2019stop} argues that post hoc explanations tend to be unfaithful to the model to the extent that their usefulness is severely compromised. While this line of work demonstrates different ways in which explanations could potentially induce inaccuracies and biases in real world applications, they do not focus on systematic benchmarking of post hoc explanation methods which is the main goal of our work.

\section{Synthetic Dataset} \label{app:synthetic_data}
Our proposed data generation process described in Algorithm \ref{alg:dgp} is designed to encapsulate arbitrary feature dependencies, unambiguous definitions of local neighborhoods, and clear descriptions of feature influences.
In our algorithm \ref{alg:dgp}, the local neighborhood of a sample is controlled by its cluster membership $k$, while the degree of feature dependency can be easily controlled by the user setting $\Sigma_k$ to desired values.
The default value of $\Sigma_k =\mathbf{I}$, where $\mathbf{I}$ is the identity matrix, indicating that all features are independent of one another.  
The elements of the true underlying weight vector $w_k$ are sampled from a uniform distribution: $w_k \sim \mathcal{U}(l,u)$. 
We set $l=-1$ and $u=1$.
The masking vectors $m_k$ with elements $m_{k,j}$ are generated by a Bernoulli distribution: $m_{k,j} \sim \mathcal{B}(p)$, where the parameter $p$ controls the ground-truth explanation sparsity. 
We set $p=0.25$.
Further, the cluster centers are chosen as follows: when the number of features $d$ is smaller than or equal to the number of clusters $K$, then we set the first cluster center to $[1, 0,  \dots, 0]$, the second one to $[0, 1, \dots, 0]$, etc. 
We also introduce a multiplier $\kappa$ to control the distance between the cluster centers so that the cluster centers are located at $\mu_1=\kappa \cdot [1, 0,  \dots, 0]$, $\mu_2=\kappa \cdot [0, 1,  \dots, 0]$, etc.
We set $\kappa=6$. 
In general, we have observed that lower values of $\kappa$ result in more complex classification problems.
When the number of features $d$ is greater than the number of clusters $K$, we adjust the above described approach slightly: we first compute $\ell = \lfloor K/d \rfloor$, and then we compute the cluster centers for the first $d$ clusters as described above, for the next $d$ clusters we use $\mu_{d+1}= 2 \kappa \cdot [1, 0,  \dots, 0]$, $\mu_{d+2}=2\kappa \cdot [0, 1,  \dots, 0]$, etc. If $\ell=1$, we stop this procedure here, else we continue, and repeat filling up the clusters $\mu_{2d+1}= 3 \kappa \cdot [1, 0,  \dots, 0]$, $\mu_{2d+2}=3\kappa \cdot [0, 1,  \dots, 0]$, etc.

\begin{algorithm}[H]
\caption{\textsc{\textbf{SynthGauss}}}
\label{alg:dgp}
\SetKwInOut{Input}{input}\SetKwInOut{Output}{output}
\Input{~number of clusters: $K$, cluster centers: $[\mu_{1}, \mu_{2}, \dots, \mu_{K}]$, cluster variances: $[\Sigma_1, \dots, \Sigma_K]$, Weight vectors: $[w_1, \dots, w_K]$, masking vectors: $[m_1, \dots, m_K]$}
\Output{~features: $\mathbf{X}$, labels: $\mathbf{y}$}
$\mathbf{X} = \mathbf{0}_{n \times d}$ \;
$\Pi = \mathbf{0}_{n \times 1}$ \;
\For{$i= 1$:$n$} {
  $k \gets$ Cat($K$) ~~~\# Randomly picks a cluster index  \;
  $x_i \sim \mathcal{N}(\mu_{k}, \Sigma_{k})$ ~~~\# Samples Gaussian instance \;
  $\mathbf{X}[i,:] = x_i$ \;
  $\pi_{1} = \mathbb{P}(y_i = 1|\mathbf{x}_i) = \frac{\exp\big((w_k \odot m_k)^\top x_{i}\big)}{1 + \exp\big((w_k \odot m_k)^\top x_{i} \big)}$ ~~~\# Get class probability \;
  $\Pi[i] = \pi_1$ \;
}
$\tilde{\pi} = \texttt{get\_median}(\Pi)$ \;

\For{$i= 1$:$n$} {
$y_i = \mathbb{I}(\Pi[i] > \tilde{\pi})$ ~~~  \# Make sure classes are balanced \;
}
Return: $\mathbf{X}, \mathbf{y}$ \; 
\end{algorithm}

\begin{theorem}
If a given dataset encapsulates the properties of feature independence, unambiguous and well-separated local neighborhoods, and a unique ground truth explanation for each local neighborhood, the most accurate model trained on such a dataset will adhere to the unique ground truth explanation of each local neighborhood.
\end{theorem}
\emph{Proof:} Before we describe the proof, we begin by first formalizing the properties of feature independence, unambiguous and well-separated local neighborhoods, and a unique ground truth explanation for each local neighborhood using some notation. Let $A = [a_1, a_2, \cdots a_d]$ denote the vector of input features in a given dataset $D$, and let the instances in the dataset $D$ be separated into $K$ clusters (local neighborhoods) based on their proximity.

Feature independence: The dataset $D$ satisfies feature independence if $P(a_i | a_j) = P(a_i)$  $\forall i, j \in \{1, 2, \cdots d\} $ and $ i \neq j$. 

Unambiguous and well-separated local neighborhoods: The dataset $D$ is said to constitute unambiguous and well-separated local neighborhoods if $dist(x_{i,j}, x_{p,q}) \gg dist(x_{i,j}, x_{i,l})$ $\forall i, p \in \{1, 2 \cdots K\}$ where $x_{i,j}$ and $x_{i,l}$ are the $j^{th}$ and $l^{th}$ instances of some cluster $i$, $x_{p,q}$ is the $q^{th}$ instance of cluster $p$ and $p \neq i$. The above implies that the distances between points in different clusters should be significantly higher than the distances between points in the same cluster.

\looseness=-1
Unique ground truth explanation for each local neighborhood: The dataset $D$ is said to comprise of unique ground truth explanations for each local neighborhood if the ground truth labels of instances in each cluster $k \in \{1, 2 \cdots K\}$ are generated as a function of a subset of features $A_k \subseteq A$ where there is a clear relative ordering among features in $A_k$ which is captured by the weight vector $w_k$, and features in $A$ are completely independent of each other. [Note that the influential feature set $A_k$ corresponding to the cluster $k$ is also captured using the mask vector $m_k$ (See Section~\ref{sec:overview}) where $m_{k,i} = 1$ if the $i^{th}$ feature is in set $A_k$]. This would imply that there is a clear description of the top-$T$ features (and their relative ordering) where $T \leq |A_k|$ influence the ground truth labeling process. 

With the above information in place, let us now consider a model $\mathcal{M}$ which is trained on the dataset $D$ and achieves highest possible accuracy on it. To show that this model indeed adheres to the unique ground truth explanations of each of the local neighborhoods (clusters), we adopt the strategy of proof by contradiction. To this end, we begin with the assumption that the model $\mathcal{M}$ does not adhere to the unique ground truth explanation of some local neighborhood $k \in \{1, 2, \cdots K\}$ i.e., the top-$T$ features leveraged by model $\mathcal{M}$ for instances in cluster $k$ (denoted by $\mathcal{M}^{k}_T$) do not exactly match the top-$T$ features leveraged by the ground truth labeling process where $T \leq |A_k|$. 
This happens either when there is a mismatch in the relative ordering among the top-$T$ features used by the model $\mathcal{M}$ and the ground truth labeling process (or) when there is at least one feature $a$ that appears among the top-$T$ features used by the model but not the ground truth labeling process. In either case, we can construct another model $\mathcal{M}'$ such that the top-$T$ features used by $\mathcal{M}'$ match the top-$T$ features of the ground truth labeling process exactly, and the accuracy of $\mathcal{M}'$ will be higher than that of $\mathcal{M}$. 
This contradicts the assumption that the model $\mathcal{M}$ has the highest possible accuracy, thus demonstrating that the model with highest possible accuracy would adhere to the unique ground truth explanation of each local neighborhood. 

\section{Benchmarking Analysis}
\label{app:exp}

\subsection{Real-World Datasets}
\label{app:datasets}


In addition to the synthetic dataset, the \name library includes seven real-world benchmark datasets from high-stakes domains. Table~\ref{tab:dataset} provides a summary of the real-world datasets currently included in \name. Our library implements multiple data split strategies and allows users to customize the percentages of train-test splits. To this end, if a given dataset comes with a pre-determined train and test splits, \name loads the training and testing dataloaders from those pre-determined splits. Otherwise, \name's dataloader divides the entire dataset randomly into train (80\%) and test (20\%). Next, we detail the covariates $\mathbf{x}$ and labels $\mathbf{y}$ for each dataset.

\textbf{German Credit}~\citep{Dua:2019} The dataset comprises of demographic (age, gender), personal (marital status), and financial (income, credit duration) features from 1,000 credit applicants, where they are categorized into good vs. bad customer depending on their credit risk.

\textbf{HELOC}~\citep{HELOC} 
The dataset comprises of financial (e.g., total number of trades, average credit months in file) attributes from anonymized applications submitted by 9,871 real homeowners. A HELOC (Home Equity Line of Credit) is a line of credit typically offered by a bank as a percentage of home equity. The fundamental task is to use the information about the applicant in their credit report to predict whether they will repay their HELOC account within 2 years. 

\textbf{Adult Income}~\citep{yeh2009comparisons} The dataset contains demographic (e.g., age, race, and gender), education (degree), employment (occupation, hours-per week), personal (marital status, relationship), and financial (capital gain/loss) features for 45,222 individuals. The task is to predict whether an individual’s income exceeds \$50K per year vs. not.

\textbf{COMPAS}~\citep{jordan2015effect} The dataset has criminal records and demographics features for 6,172 defendants released on bail at U.S state courts during 1990-2009. The task is to classify defendants into bail (i.e., unlikely to commit a violent crime if released) vs. no bail (i.e., likely to commit a violent crime).

\textbf{Give Me Some Credit}~\citep{GiveMeCredit} The dataset incorporates demographic (age), personal (number of dependents), and financial (e.g., monthly income, debt ratio, etc.) features for 102,209 individuals. The task is to predict the probability whether or not a customer will experience financial distress in the next two years. The aim of the dataset is to build models that customers can use to the best financial decisions.

We also include the \textbf{Pima-Indians Diabetes}~\citep{smith1988using} and \textbf{Framingham Heart Study}~\citep{Framingh21} datasets (see Appendix~\ref{sec:rebuttal}).


\subsection{Explanation Methods}
\name provides implementations for six explanation methods:  \Grads~\citep{simonyan2013saliency}, \IntGrad~\citep{sundararajan2017axiomatic}, \SmoothGrad~\citep{smilkov2017smoothgrad}, \GradtimesInput~\citep{shrikumar2017learning}, \LIME~\citep{ribeiro16:kdd}, and \SHAP~\citep{lundberg2017unified}. Table~\ref{tab:explanation} summarizes how these methods differ along two axes:  whether each method requires ``White-box Access'' to model gradients and whether each method learns a local approximation model to generate explanations.

\begin{minipage}[c]{0.99\textwidth}
    \centering\small
    \renewcommand{\arraystretch}{0.9}
    \setlength{\tabcolsep}{1.8pt}
    \captionof{table}{
        Summary of currently available feature attribution methods in \name. “Learning?” denotes whether learning/training procedures are required to generate explanations and "White-box Access?" denotes if the access to the internals of the model (e.g., gradients) are needed. 
    }    
    {\begin{tabular}{lcccc}
    \toprule
    {Method} & {Learning?} & {White-box Access?} \\ 
    \toprule
    \begin{tabular}[l]{@{}l@{}}{Random}\\{VanillaGrad}\\{IntegratedGrad}\\{\GradtimesInput}\\{\SmoothGrad}\\{\SHAP}\\{\LIME}
    \end{tabular} &
    \begin{tabular}[c]{@{}c@{}}
        \xmark\\
        \xmark\\
        \xmark\\
        \xmark\\
        \xmark\\
        \cmark\\
        \cmark\\
    \end{tabular} &
    \begin{tabular}[c]{@{}c@{}}
        \xmark\\
        \cmark\\
        \cmark\\
        \cmark\\
        \cmark\\
        \xmark\\
        \xmark\\
    \end{tabular} \\
    \bottomrule \\
    \end{tabular}}
    \label{tab:explanation}
\end{minipage}

\looseness=-1
\xhdr{Implementations}  We used existing public implementations of all explanation methods in our experiments.  We used the following \texttt{captum} software package classes: i) \texttt{captum.attr.Saliency} for \Grads; ii) \texttt{captum.attr.IntegratedGradients} for \IntGrad; iii) \texttt{captum.attr.NoiseTunnel} and \texttt{captum.attr.Saliency} for \SmoothGrad; iv) \texttt{captum.attr.InputXGradient} for \GradtimesInput; and v) \texttt{captum.attr.KernelShap} for \SHAP. Finally, we use the authors' \href{https://github.com/marcotcr/lime}{python package} for \LIME.

\subsection{Hyperparameter details}
\label{app:parameter}

\name uses default hyperparameter settings for all explanation methods following the authors' guidelines. 
Every explanation method has a corresponding parameter dictionary that stores all the default parameter values.
Below, we detail the hyperparameters of individual explanation methods used in our experiments. Seed is set to 0 for all explanation methods to allow reproducibility.
        
\looseness=-1
\textbf{a) params\_lime} {=} \{`\textit{seed}': 0, `\textit{lime\_mode}': `tabular', `\textit{sample\_around\_instance}': True, `\textit{kernel\_ width}': 0.75, `\textit{n\_samples}': 1000, `\textit{discretize\_continuous}': False, `\textit{std}': 0.1\}

\emph{Description.} The option `tabular' indicates that we wish to compute explanations on a tabular data set. The option `lime\_sample\_around\_instance' makes sure that the sampling is conducted in a local neighborhood around the point that we wish to explain. The `lime\_discretize\_continuous' option ensures that continuous variables are kept continuous, and are not discretized. This parameter sets the `lime\_standard\_ deviation' of the Gaussian random variable that is used to sample around the instance that we wish to explain. We set this to a small value, ensuring that we in fact sample from a local neighborhood.

\textbf{b) params\_shap} {=} \{`\textit{seed}': 0, `\textit{n\_samples}': 500, `\textit{model\_impl}': `torch'\}

\emph{Description.} The parameter `shap\_n\_samples' controls the number of samples of the original model used to train the surrogate SHAP model, while the parameter `model\_impl' toggles between models from `torch' and `sklearn'.

\textbf{c) params\_grads} {=} \{`\textit{absolute\_value}': False\}

\emph{Description.} The parameter `grad\_absolute\_value' controls whether the absolute value of each element in the explanation vector should be taken or not. 

\textbf{d) params\_sg} = \{`\textit{n\_samples}': 500, `\textit{standard\_deviation}': 0.1\}

\emph{Description.} The parameter `sg\_n\_samples' sets the number of samples used in the Gaussian random variable to smooth the gradient, while `sg\_standard\_deviation' determines the size of the local neighborhood the Gaussian random variables are sampled from. We use a small value, ensuring that we sample from a local neighborhood.

\textbf{e) params\_ig} = \{`\textit{method}': `gausslegendre', `\textit{multiply\_by\_inputs}': False\}

Further, we parameterized our data-generating process as follows.

\textbf{f) params\_gauss} = \{`\textit{n\_samples}': 1000, `\textit{dim}':20, `\textit{n\_clusters}': 10, `\textit{distance\_to\_center}': 6, `\textit{test\_size}': 0.25, `\textit{upper\_weight}': 1, `\textit{lower\_weight}': -1, `\textit{seed}': 564, `\textit{sigma}': None, `\textit{sparsity}': 0.25\}

\emph{Description.} The above parameters can be matched with the parameters from the data generating process described in Appendix \ref{app:synthetic_data}: in particular, `sparsity'$=p$, `upper\_weight' $=u$, `lower\_weight' $=l$, `dim'$=d$, `n\_clusters' $=K$, `distance\_to\_center' $=\kappa$ and `sigma'=None implies that the identity matrix is used, i.e., $\Sigma_{k}=\mathbf{I}$.

In addition, for a given instance $\mathbf{x}$, some evaluation metrics leverage a perturbation class to generate perturbed samples $\mathbf{x}'$. We have a parameterized version of this perturbation class in \name with the following default parameters:

\textbf{g) params\_perturb} = \{`\textit{perturbation\_mean}' : 0.0, `perturbation\_std' : 0.05, `\textit{perturbation\_ flip\_percentage}' : 0.03\}

Finally, we dynamically infer the top-$k$ value from the input parameter $k$ and the total number of features, $d$, as follows: top-$k=k$ for $\{ k \in \mathbb{Z} \mid 1 \leq k \leq d \}$; top-$k=d$ for $k=-1$; and top-$k=\lceil k \times d \rceil$ for $\{k \in \mathbb{R} \mid 0 < k < 1\}$.

For predictive faithfulness, we set $\sigma{=}0.1$, and sample 100 perturbations. The flip percentage is set to $\sigma\sqrt{2/\pi}$ in all cases, which leads to the same expected magnitude of change between continuous and discrete variables when drawing from a Normal distribution. For stability, we set $\sigma{=}10^{-5}$, and sample 1000 points before selecting a subset of 100 samples that share the original input's prediction.

The $p$ norm for all stability metrics is set to 2. Seed is set to the special case of -1 for our evaluation metrics, which assumes the index of the specific test instance being evaluated as the seed being used. All default hyperparameters used in our experiments can also be found in the \texttt{experiment\_config.json} file of our \href{https://github.com/AI4LIFE-GROUP/OpenXAI}{OpenXAI GitHub repository}.

\subsection{Model details}
\label{app:model}

Our current release of \name has two pre-trained models: i) a logistic regression model and ii) a deep neural network model for all datasets. To support systematic, reproducible, and efficient benchmarking of post hoc explanation methods, we provide the model weights for both models trained on all eight datasets in our pipeline. For neural network models, we use two fully connected hidden layers with 100 nodes in each layer, with ReLU activation functions and an output softmax layer for all datasets. We train both models for 100 epochs using an Adam optimizer with a learning rate of $\eta{=}0.001$. We use min-max scaling to normalize train/test splits, along with dataset-specific batch sizes/positive class weights (to balance the loss function). The best epoch in terms of test accuracy is selected, excluding an initial warmup phase of 5 epochs, and provided that dataset-specific bounds are not exceeded for the proportion of the majority class predicted. Next, we show the performance of both models on all eight datasets.

\begin{table}[h]
    \centering\small
    \renewcommand{\arraystretch}{0.9}
    \setlength{\tabcolsep}{1.5pt}
    \caption{
        \textbf{Results of the machine learning models trained on eight datasets.} Shown are the test accuracies of LR and ANN models. Values corresponding to best performance are bolded.
    }
    {\begin{tabular}{lcc}
    \toprule
    {Dataset} & {LR} & {ANN} \\
    \toprule
    \begin{tabular}[l]{@{}l@{}}{Synthetic Data}\\{German~Credit}\\{HELOC}\\{COMPAS}\\{Adult~Income}\\{Give Me Some Credit}\\{Pima-Indians Diabetes}\\{Framingham Heart Study}
    \end{tabular} &
    \begin{tabular}[c]{@{}c@{}}
    {81.76\%}\\
    \textbf{64.00\%}\\
    {72.35\%}\\
    \textbf{85.26\%}\\
    {83.25\%}\\
    \textbf{93.79\%}\\
    {69.48\%}\\
    {80.19\%}\\
    \end{tabular} &
    \begin{tabular}[c]{@{}c@{}}
    \textbf{90.32\%}\\
    {61.50\%}\\
    \textbf{73.92\%}\\
    {85.02\%}\\
    \textbf{84.62\%}\\
    {93.43\%}\\
    \textbf{77.92\%}\\
    \textbf{80.87}\%\\
    \end{tabular} \\
    \bottomrule
    \end{tabular}}
    \label{tab:performance}
\end{table}

\subsection{Results for Pima-Indians Diabetes and Framingham Heart Study datasets}
\label{sec:rebuttal}

We include two new datasets from the healthcare domain and benchmark state-of-the-art explanation methods on these datasets. More specifically, we include the \textbf{Pima-Indians Diabetes}~\citep{smith1988using} and \textbf{Framingham Heart Study}~\citep{Framingh21} datasets both of which have been utilized in recent XAI research. New results with these datasets are shown below. We observe similar insights with these new datasets: predictive faithfulness results are shown in Tables ~\ref{tab:pima_faith_lr}-\ref{tab:heart_faith_ann} below); stability results are shown in Tables~\ref{tab:pima_stab_lr}-\ref{tab:heart_stab_ann} below.

\begin{table}[h]
    \centering\small
    \renewcommand{\arraystretch}{0.9}
    \setlength{\tabcolsep}{1.5pt}
    \caption{
        \textbf{Ground-truth and predictive faithfulness results on the Pima Indians Diabetes dataset for all explanation methods with LR model.} Shown are average and standard error metric values computed across 1000 test instances. $\uparrow$ indicates that higher values are better, and $\downarrow$ indicates that lower values are better. Values corresponding to best performance are bolded.
    }
    {\begin{tabular}{lcccccccc}
    \toprule
    {{Method}} & {PRA~($\uparrow$)} & {RC~($\uparrow$)} & {FA~($\uparrow$)} & {RA~($\uparrow$)} & {SA~($\uparrow$)} & {SRA~($\uparrow$)} & {PGU~($\downarrow$)} & {PGI~($\uparrow$)} \\
    \toprule
    \begin{tabular}[l]{@{}l@{}}{Random}\\{VanillaGrad}\\{IntegratedGrad}\\{\GradtimesInput}\\{\SmoothGrad}\\{\SHAP}\\{\LIME}
    \end{tabular} &
	\begin{tabular}[c]{@{}c@{}}
	{0.470}\std{0.01}\\
	\textbf{1.000}\std{0.00}\\
	\textbf{1.000}\std{0.00}\\
	{0.492}\std{0.01}\\
	\textbf{1.000}\std{0.00}\\
	{0.488}\std{0.01}\\
	{0.980}\std{0.00}\\
	\end{tabular} &
	\begin{tabular}[c]{@{}c@{}}
	{-0.076}\std{0.03}\\
	\textbf{1.000}\std{0.00}\\
	\textbf{1.000}\std{0.00}\\
	{-0.047}\std{0.03}\\
	\textbf{1.000}\std{0.00}\\
	{-0.047}\std{0.03}\\
	{0.986}\std{0.00}\\
	\end{tabular} &
	\begin{tabular}[c]{@{}c@{}}
	{0.154}\std{0.02}\\
	\textbf{1.000}\std{0.00}\\
	\textbf{1.000}\std{0.00}\\
	{0.195}\std{0.02}\\
	\textbf{1.000}\std{0.00}\\
	{0.195}\std{0.02}\\
	{0.985}\std{0.00}\\
	\end{tabular} &
	\begin{tabular}[c]{@{}c@{}}
	{0.109}\std{0.02}\\
	\textbf{1.000}\std{0.00}\\
	\textbf{1.000}\std{0.00}\\
	{0.133}\std{0.02}\\
	\textbf{1.000}\std{0.00}\\
	{0.135}\std{0.02}\\
	{0.985}\std{0.00}\\
	\end{tabular} &
	\begin{tabular}[c]{@{}c@{}}
	{0.075}\std{0.02}\\
	\textbf{1.000}\std{0.00}\\
	\textbf{1.000}\std{0.00}\\
	{0.195}\std{0.02}\\
	\textbf{1.000}\std{0.00}\\
	{0.195}\std{0.02}\\
	{0.985}\std{0.00}\\
	\end{tabular} &
	\begin{tabular}[c]{@{}c@{}}
	{0.054}\std{0.01}\\
	\textbf{1.000}\std{0.00}\\
	\textbf{1.000}\std{0.00}\\
	{0.133}\std{0.02}\\
	\textbf{1.000}\std{0.00}\\
	{0.135}\std{0.02}\\
	{0.985}\std{0.00}\\
	\end{tabular} &
	\begin{tabular}[c]{@{}c@{}}
	{0.031}\std{0.00}\\
	\textbf{0.026}\std{0.00}\\
	\textbf{0.026}\std{0.00}\\
	{0.030}\std{0.00}\\
	\textbf{0.026}\std{0.00}\\
	{0.030}\std{0.00}\\
	\textbf{0.026}\std{0.00}\\
	\end{tabular} &
	\begin{tabular}[c]{@{}c@{}}
	{0.013}\std{0.00}\\
	\textbf{0.021}\std{0.00}\\
	\textbf{0.021}\std{0.00}\\
	{0.015}\std{0.00}\\
	\textbf{0.021}\std{0.00}\\
	{0.015}\std{0.00}\\
	\textbf{0.021}\std{0.00}\\
	\end{tabular}\\
    \bottomrule
    \end{tabular}}
    \label{tab:pima_faith_lr}
\end{table}
\begin{table}[h]
    \centering\small
    \renewcommand{\arraystretch}{0.9}
    \setlength{\tabcolsep}{1.5pt}
    \caption{
        \textbf{Ground-truth and predictive faithfulness results on the Framingham Heart dataset for all explanation methods with LR model.} Shown are average and standard error metric values computed across 1000 test instances. $\uparrow$ indicates that higher values are better, and $\downarrow$ indicates that lower values are better. Values corresponding to best performance are bolded.
    }
    {\begin{tabular}{lcccccccc}
    \toprule
    {{Method}} & {PRA~($\uparrow$)} & {RC~($\uparrow$)} & {FA~($\uparrow$)} & {RA~($\uparrow$)} & {SA~($\uparrow$)} & {SRA~($\uparrow$)} & {PGU~($\downarrow$)} & {PGI~($\uparrow$)} \\
    \toprule
    \begin{tabular}[l]{@{}l@{}}{Random}\\{VanillaGrad}\\{IntegratedGrad}\\{\GradtimesInput}\\{\SmoothGrad}\\{\SHAP}\\{\LIME}
    \end{tabular} &
	\begin{tabular}[c]{@{}c@{}}
	{0.495}\std{0.00}\\
	\textbf{1.000}\std{0.00}\\
	\textbf{1.000}\std{0.00}\\
	{0.619}\std{0.00}\\
	\textbf{1.000}\std{0.00}\\
	{0.573}\std{0.00}\\
	{0.978}\std{0.00}\\
	\end{tabular} &
	\begin{tabular}[c]{@{}c@{}}
	{-0.013}\std{0.01}\\
	\textbf{1.000}\std{0.00}\\
	\textbf{1.000}\std{0.00}\\
	{0.280}\std{0.01}\\
	\textbf{1.000}\std{0.00}\\
	{0.208}\std{0.01}\\
	{0.991}\std{0.00}\\
	\end{tabular} &
	\begin{tabular}[c]{@{}c@{}}
	{0.168}\std{0.01}\\
	\textbf{1.000}\std{0.00}\\
	\textbf{1.000}\std{0.00}\\
	{0.472}\std{0.01}\\
	\textbf{1.000}\std{0.00}\\
	{0.475}\std{0.01}\\
	{0.971}\std{0.00}\\
	\end{tabular} &
	\begin{tabular}[c]{@{}c@{}}
	{0.067}\std{0.01}\\
	\textbf{1.000}\std{0.00}\\
	\textbf{1.000}\std{0.00}\\
	{0.313}\std{0.01}\\
	\textbf{1.000}\std{0.00}\\
	{0.321}\std{0.01}\\
	{0.951}\std{0.00}\\
	\end{tabular} &
	\begin{tabular}[c]{@{}c@{}}
	{0.090}\std{0.01}\\
	\textbf{1.000}\std{0.00}\\
	\textbf{1.000}\std{0.00}\\
	{0.472}\std{0.01}\\
	\textbf{1.000}\std{0.00}\\
	{0.475}\std{0.01}\\
	{0.971}\std{0.00}\\
	\end{tabular} &
	\begin{tabular}[c]{@{}c@{}}
	{0.040}\std{0.00}\\
	\textbf{1.000}\std{0.00}\\
	\textbf{1.000}\std{0.00}\\
	{0.313}\std{0.01}\\
	\textbf{1.000}\std{0.00}\\
	{0.321}\std{0.01}\\
	{0.951}\std{0.00}\\
	\end{tabular} &
	\begin{tabular}[c]{@{}c@{}}
	{0.086}\std{0.00}\\
	\textbf{0.074}\std{0.00}\\
	\textbf{0.074}\std{0.00}\\
	{0.089}\std{0.00}\\
	\textbf{0.074}\std{0.00}\\
	{0.089}\std{0.00}\\
	\textbf{0.074}\std{0.00}\\
	\end{tabular} &
	\begin{tabular}[c]{@{}c@{}}
	{0.025}\std{0.00}\\
	\textbf{0.042}\std{0.00}\\
	\textbf{0.042}\std{0.00}\\
	{0.029}\std{0.00}\\
	\textbf{0.042}\std{0.00}\\
	{0.029}\std{0.00}\\
	\textbf{0.042}\std{0.00}\\
	\end{tabular}\\
    \bottomrule
    \end{tabular}}
    \label{tab:heart_faith_lr}
\end{table}

\vspace{20pt}
\begin{minipage}[b]{0.48\textwidth}
	\centering\small
    \renewcommand{\arraystretch}{0.9}
    \setlength{\tabcolsep}{2.0pt}
    \captionof{table}{
        \textbf{Predictive faithfulness results on the Pima Indians Diabetes dataset for all explanation methods with ANN model.} Shown are average and standard error metric values computed across 1000 test instances. $\uparrow$ indicates that higher values are better, and $\downarrow$ indicates that lower values are better. Values corresponding to best performance are bolded.
    }
    {\begin{tabular}{lcc}
    \toprule
    {{Method}} & {PGU~($\downarrow$)} & {PGI~($\uparrow$)} \\
    \toprule
    \begin{tabular}[l]{@{}l@{}}{Random}\\{VanillaGrad}\\{IntegratedGrad}\\{\GradtimesInput}\\{\SmoothGrad}\\{\SHAP}\\{\LIME}
    \end{tabular} &
	\begin{tabular}[c]{@{}c@{}}
	{0.084}\std{0.008}\\
	{0.055}\std{0.006}\\
	{0.055}\std{0.006}\\
	{0.058}\std{0.006}\\
	\textbf{0.054}\std{0.006}\\
	{0.069}\std{0.007}\\
	\textbf{0.054}\std{0.006}\\
	\end{tabular} &
	\begin{tabular}[c]{@{}c@{}}
	{0.034}\std{0.004}\\
	\textbf{0.074}\std{0.007}\\
	\textbf{0.074}\std{0.007}\\
	{0.070}\std{0.007}\\
	\textbf{0.074}\std{0.007}\\
	{0.056}\std{0.006}\\
	\textbf{0.074}\std{0.007}\\
	\end{tabular}\\
    \bottomrule
    \end{tabular}}
    \label{tab:pima_faith_ann}
\end{minipage}
\hfill
\begin{minipage}[b]{0.48\textwidth}
	\centering\small
    \renewcommand{\arraystretch}{0.9}
    \setlength{\tabcolsep}{2.0pt}
    \captionof{table}{
        \textbf{Predictive faithfulness results on the Framingham Heart dataset for all explanation methods with ANN model.} Shown are average and standard error metric values computed across 1000 test instances. $\uparrow$ indicates that higher values are better, and $\downarrow$ indicates that lower values are better. Values corresponding to best performance are bolded.
    }
    {\begin{tabular}{lcc}
    \toprule
    {{Method}} & {PGU~($\downarrow$)} & {PGI~($\uparrow$)} \\
    \toprule
    \begin{tabular}[l]{@{}l@{}}{Random}\\{VanillaGrad}\\{IntegratedGrad}\\{\GradtimesInput}\\{\SmoothGrad}\\{\SHAP}\\{\LIME}
    \end{tabular} &
	\begin{tabular}[c]{@{}c@{}}
	{0.130}\std{0.005}\\
	\textbf{0.110}\std{0.005}\\
	{0.112}\std{0.005}\\
	{0.118}\std{0.005}\\
	\textbf{0.110}\std{0.005}\\
	{0.121}\std{0.005}\\
	\textbf{0.110}\std{0.005}\\
	\end{tabular} &
	\begin{tabular}[c]{@{}c@{}}
	{0.047}\std{0.003}\\
	\textbf{0.089}\std{0.004}\\
	{0.085}\std{0.004}\\
	{0.079}\std{0.004}\\
	{0.088}\std{0.004}\\
	{0.073}\std{0.004}\\
	{0.088}\std{0.004}\\
	\end{tabular}\\
    \bottomrule
    \end{tabular}}
    \label{tab:heart_faith_ann}
\end{minipage}

\begin{minipage}[b]{0.48\textwidth}
	\centering\small
    \renewcommand{\arraystretch}{0.9}
    \setlength{\tabcolsep}{2.0pt}
    \captionof{table}{
        \textbf{Stability results on the Pima Indians Diabetes dataset for all explanation methods with LR model.} Shown are log average and standard error metric values computed across 1000 test instances. $\uparrow$ indicates that higher values are better, and $\downarrow$ indicates that lower values are better. Values corresponding to best performance are bolded.
    }
    {\begin{tabular}{lcc}
    \toprule
    {{Method}} & {RIS~($\downarrow$)} & {ROS~($\downarrow$)} \\
    \toprule
    \begin{tabular}[l]{@{}l@{}}{Random}\\{VanillaGrad}\\{IntegratedGrad}\\{\GradtimesInput}\\{\SmoothGrad}\\{\SHAP}\\{\LIME}
    \end{tabular} &
	\begin{tabular}[c]{@{}c@{}}
	{11.86}\std{0.02}\\
	{-1.01}\std{0.06}\\
	\textbf{-2.12}\std{0.06}\\
	{0.70}\std{0.02}\\
	{6.81}\std{0.06}\\
	{6.72}\std{0.04}\\
	{7.17}\std{0.02}\\
	\end{tabular} &
	\begin{tabular}[c]{@{}c@{}}
	{13.11}\std{0.04}\\
	{-2.04}\std{0.08}\\
	\textbf{-3.20}\std{0.08}\\
	{2.75}\std{0.10}\\
	{7.52}\std{0.08}\\
	{7.77}\std{0.08}\\
	{8.37}\std{0.05}\\
	\end{tabular}\\
    \bottomrule
    \end{tabular}}
    \label{tab:pima_stab_lr}
\end{minipage}
\hfill
\begin{minipage}[b]{0.48\textwidth}
	\centering\small
    \renewcommand{\arraystretch}{0.9}
    \setlength{\tabcolsep}{2.0pt}
    \captionof{table}{
        \textbf{Stability results on the Framingham Heart dataset for all explanation methods with LR model.} Shown are log average and standard error metric values computed across 1000 test instances. $\uparrow$ indicates that higher values are better, and $\downarrow$ indicates that lower values are better. Values corresponding to best performance are bolded.
    }
    {\begin{tabular}{lcc}
    \toprule
    {{Method}} & {RIS~($\downarrow$)} & {ROS~($\downarrow$)} \\
    \toprule
    \begin{tabular}[l]{@{}l@{}}{Random}\\{VanillaGrad}\\{IntegratedGrad}\\{\GradtimesInput}\\{\SmoothGrad}\\{\SHAP}\\{\LIME}
    \end{tabular} &
	\begin{tabular}[c]{@{}c@{}}
	{11.76}\std{0.01}\\
	{1.25}\std{0.02}\\
	\textbf{0.43}\std{0.01}\\
	{1.41}\std{0.01}\\
	{8.59}\std{0.01}\\
	{9.64}\std{0.01}\\
	{8.95}\std{0.01}\\
	\end{tabular} &
	\begin{tabular}[c]{@{}c@{}}
	{15.47}\std{0.02}\\
	{0.80}\std{0.04}\\
	\textbf{-0.30}\std{0.02}\\
	{3.55}\std{0.08}\\
	{11.22}\std{0.03}\\
	{11.73}\std{0.02}\\
	{10.67}\std{0.02}\\
	\end{tabular}\\
    \bottomrule
    \end{tabular}}
    \label{tab:heart_stab_lr}
\end{minipage}

\begin{minipage}[b]{0.48\textwidth}
	\centering\small
    \renewcommand{\arraystretch}{0.9}
    \setlength{\tabcolsep}{2.0pt}
    \captionof{table}{
        \textbf{Stability results on the Pima Indians Diabetes dataset for all explanation methods with ANN model.} Shown are log average and standard error metric values computed across 1000 test instances. $\uparrow$ indicates that higher values are better, and $\downarrow$ indicates that lower values are better. Values corresponding to best performance are bolded.
    }
    {\begin{tabular}{lccc}
    \toprule
    {{Method}} & {RIS~($\downarrow$)} & {RRS~($\downarrow$)} & {ROS~($\downarrow$)} \\
    \toprule
    \begin{tabular}[l]{@{}l@{}}{Random}\\{VanillaGrad}\\{IntegratedGrad}\\{\GradtimesInput}\\{\SmoothGrad}\\{\SHAP}\\{\LIME}
    \end{tabular} &
	\begin{tabular}[c]{@{}c@{}}
	{11.86}\std{0.02}\\
	{4.37}\std{0.70}\\
	\textbf{1.95}\std{0.19}\\
	{4.59}\std{0.76}\\
	{10.85}\std{0.09}\\
	{10.31}\std{0.04}\\
	{10.83}\std{0.11}\\
	\end{tabular} &
	\begin{tabular}[c]{@{}c@{}}
	{11.90}\std{0.01}\\
	{4.54}\std{0.76}\\
	\textbf{1.93}\std{0.22}\\
	{4.77}\std{0.80}\\
	{10.93}\std{0.08}\\
	{10.36}\std{0.04}\\
	{11.02}\std{0.12}\\
	\end{tabular} &
	\begin{tabular}[c]{@{}c@{}}
	{15.06}\std{0.10}\\
	{8.42}\std{0.99}\\
	\textbf{2.91}\std{0.15}\\
	{8.72}\std{0.99}\\
	{13.83}\std{0.10}\\
	{13.36}\std{0.16}\\
	{13.94}\std{0.15}\\
	\end{tabular}\\
    \bottomrule
    \end{tabular}}
    \label{tab:pima_stab_ann}
\end{minipage}
\hfill
\begin{minipage}[b]{0.48\textwidth}
	\centering\small
    \renewcommand{\arraystretch}{0.9}
    \setlength{\tabcolsep}{2.0pt}
    \captionof{table}{
        \textbf{Stability results on the Framingham Heart dataset for all explanation methods with ANN model.} Shown are log average and standard error metric values computed across 1000 test instances. $\uparrow$ indicates that higher values are better, and $\downarrow$ indicates that lower values are better. Values corresponding to best performance are bolded.
    }
    {\begin{tabular}{lccc}
    \toprule
    {{Method}} & {RIS~($\downarrow$)} & {RRS~($\downarrow$)} & {ROS~($\downarrow$)} \\
    \toprule
    \begin{tabular}[l]{@{}l@{}}{Random}\\{VanillaGrad}\\{IntegratedGrad}\\{\GradtimesInput}\\{\SmoothGrad}\\{\SHAP}\\{\LIME}
    \end{tabular} &
	\begin{tabular}[c]{@{}c@{}}
	{11.76}\std{0.01}\\
	{5.06}\std{0.17}\\
	\textbf{4.11}\std{0.17}\\
	{5.09}\std{0.18}\\
	{26.14}\std{0.30}\\
	{16.77}\std{0.32}\\
	{26.56}\std{0.48}\\
	\end{tabular} &
	\begin{tabular}[c]{@{}c@{}}
	{11.25}\std{0.01}\\
	{4.19}\std{0.16}\\
	\textbf{3.23}\std{0.21}\\
	{4.21}\std{0.17}\\
	{25.94}\std{0.29}\\
	{16.41}\std{0.31}\\
	{26.52}\std{0.47}\\
	\end{tabular} &
	\begin{tabular}[c]{@{}c@{}}
	{15.21}\std{0.05}\\
	\textbf{4.50}\std{0.37}\\
	{5.34}\std{0.33}\\
	{4.87}\std{0.29}\\
	{26.77}\std{0.31}\\
	{19.49}\std{0.29}\\
	{27.43}\std{0.47}\\
	\end{tabular}\\
    \bottomrule
    \end{tabular}}
    \label{tab:heart_stab_ann}
\end{minipage}

\subsection{Remaining Results on LR models}
\label{app:result_lr}

\begin{table}[h]
    \centering\small
    \renewcommand{\arraystretch}{0.9}
    \setlength{\tabcolsep}{1.5pt}
    \caption{
        \textbf{Ground-truth and predictive faithfulness results on the Synthetic dataset for all explanation methods with LR model.} Shown are average and standard error metric values computed across 1000 test instances. $\uparrow$ indicates that higher values are better, and $\downarrow$ indicates that lower values are better. Values corresponding to best performance are bolded.
    }
    {\begin{tabular}{lcccccccc}
    \toprule
    {{Method}} & {PRA~($\uparrow$)} & {RC~($\uparrow$)} & {FA~($\uparrow$)} & {RA~($\uparrow$)} & {SA~($\uparrow$)} & {SRA~($\uparrow$)} & {PGU~($\downarrow$)} & {PGI~($\uparrow$)} \\
    \toprule
    \begin{tabular}[l]{@{}l@{}}{Random}\\{VanillaGrad}\\{IntegratedGrad}\\{\GradtimesInput}\\{\SmoothGrad}\\{\SHAP}\\{\LIME}
    \end{tabular} &
	\begin{tabular}[c]{@{}c@{}}
	{0.503}\std{0.00}\\
	\textbf{1.000}\std{0.00}\\
	\textbf{1.000}\std{0.00}\\
	{0.877}\std{0.00}\\
	\textbf{1.000}\std{0.00}\\
	{0.854}\std{0.00}\\
	{0.950}\std{0.00}\\
	\end{tabular} &
	\begin{tabular}[c]{@{}c@{}}
	{0.008}\std{0.01}\\
	\textbf{1.000}\std{0.00}\\
	\textbf{1.000}\std{0.00}\\
	{0.895}\std{0.00}\\
	\textbf{1.000}\std{0.00}\\
	{0.868}\std{0.00}\\
	{0.973}\std{0.00}\\
	\end{tabular} &
	\begin{tabular}[c]{@{}c@{}}
	{0.155}\std{0.00}\\
	\textbf{1.000}\std{0.00}\\
	\textbf{1.000}\std{0.00}\\
	{0.555}\std{0.01}\\
	\textbf{1.000}\std{0.00}\\
	{0.544}\std{0.00}\\
	{0.822}\std{0.00}\\
	\end{tabular} &
	\begin{tabular}[c]{@{}c@{}}
	{0.052}\std{0.00}\\
	\textbf{1.000}\std{0.00}\\
	\textbf{1.000}\std{0.00}\\
	{0.210}\std{0.01}\\
	\textbf{1.000}\std{0.00}\\
	{0.181}\std{0.01}\\
	{0.433}\std{0.01}\\
	\end{tabular} &
	\begin{tabular}[c]{@{}c@{}}
	{0.078}\std{0.00}\\
	\textbf{1.000}\std{0.00}\\
	\textbf{1.000}\std{0.00}\\
	{0.555}\std{0.01}\\
	\textbf{1.000}\std{0.00}\\
	{0.544}\std{0.00}\\
	{0.822}\std{0.00}\\
	\end{tabular} &
	\begin{tabular}[c]{@{}c@{}}
	{0.027}\std{0.00}\\
	\textbf{1.000}\std{0.00}\\
	\textbf{1.000}\std{0.00}\\
	{0.210}\std{0.01}\\
	\textbf{1.000}\std{0.00}\\
	{0.181}\std{0.01}\\
	{0.433}\std{0.01}\\
	\end{tabular} &
	\begin{tabular}[c]{@{}c@{}}
	{0.122}\std{0.00}\\
	\textbf{0.090}\std{0.00}\\
	\textbf{0.090}\std{0.00}\\
	{0.100}\std{0.00}\\
	\textbf{0.090}\std{0.00}\\
	{0.100}\std{0.00}\\
	\textbf{0.090}\std{0.00}\\
	\end{tabular} &
	\begin{tabular}[c]{@{}c@{}}
	{0.047}\std{0.00}\\
	\textbf{0.096}\std{0.00}\\
	\textbf{0.096}\std{0.00}\\
	{0.086}\std{0.00}\\
	\textbf{0.096}\std{0.00}\\
	{0.086}\std{0.00}\\
	\textbf{0.096}\std{0.00}\\
	\end{tabular}\\
    \bottomrule
    \end{tabular}}
    \label{tab:gaussian_faith_lr}
\end{table}

\begin{table}[h]
    \centering\small
    \renewcommand{\arraystretch}{0.9}
    \setlength{\tabcolsep}{1.5pt}
    \caption{
        \textbf{Ground-truth and predictive faithfulness results on the German Credit dataset for all explanation methods with LR model.} Shown are average and standard error metric values computed across 1000 test instances. $\uparrow$ indicates that higher values are better, and $\downarrow$ indicates that lower values are better. Values corresponding to best performance are bolded.
    }
    {\begin{tabular}{lcccccccc}
    \toprule
    {{Method}} & {PRA~($\uparrow$)} & {RC~($\uparrow$)} & {FA~($\uparrow$)} & {RA~($\uparrow$)} & {SA~($\uparrow$)} & {SRA~($\uparrow$)} & {PGU~($\downarrow$)} & {PGI~($\uparrow$)} \\
    \toprule
    \begin{tabular}[l]{@{}l@{}}{Random}\\{VanillaGrad}\\{IntegratedGrad}\\{\GradtimesInput}\\{\SmoothGrad}\\{\SHAP}\\{\LIME}
    \end{tabular} &
	\begin{tabular}[c]{@{}c@{}}
	{0.500}\std{0.00}\\
	\textbf{1.000}\std{0.00}\\
	\textbf{1.000}\std{0.00}\\
	{0.465}\std{0.00}\\
	\textbf{1.000}\std{0.00}\\
	{0.505}\std{0.00}\\
	{0.977}\std{0.00}\\
	\end{tabular} &
	\begin{tabular}[c]{@{}c@{}}
	{-0.000}\std{0.01}\\
	\textbf{1.000}\std{0.00}\\
	\textbf{1.000}\std{0.00}\\
	{0.122}\std{0.01}\\
	\textbf{1.000}\std{0.00}\\
	{-0.001}\std{0.01}\\
	{0.995}\std{0.00}\\
	\end{tabular} &
	\begin{tabular}[c]{@{}c@{}}
	{0.129}\std{0.01}\\
	\textbf{1.000}\std{0.00}\\
	\textbf{1.000}\std{0.00}\\
	{0.171}\std{0.01}\\
	\textbf{1.000}\std{0.00}\\
	{0.172}\std{0.01}\\
	{0.971}\std{0.00}\\
	\end{tabular} &
	\begin{tabular}[c]{@{}c@{}}
	{0.014}\std{0.00}\\
	\textbf{1.000}\std{0.00}\\
	\textbf{1.000}\std{0.00}\\
	{0.026}\std{0.00}\\
	\textbf{1.000}\std{0.00}\\
	{0.025}\std{0.00}\\
	{0.837}\std{0.01}\\
	\end{tabular} &
	\begin{tabular}[c]{@{}c@{}}
	{0.067}\std{0.00}\\
	\textbf{1.000}\std{0.00}\\
	\textbf{1.000}\std{0.00}\\
	{0.171}\std{0.01}\\
	\textbf{1.000}\std{0.00}\\
	{0.171}\std{0.01}\\
	{0.971}\std{0.00}\\
	\end{tabular} &
	\begin{tabular}[c]{@{}c@{}}
	{0.007}\std{0.00}\\
	\textbf{1.000}\std{0.00}\\
	\textbf{1.000}\std{0.00}\\
	{0.026}\std{0.00}\\
	\textbf{1.000}\std{0.00}\\
	{0.025}\std{0.00}\\
	{0.837}\std{0.01}\\
	\end{tabular} &
	\begin{tabular}[c]{@{}c@{}}
	{0.049}\std{0.00}\\
	\textbf{0.035}\std{0.00}\\
	\textbf{0.035}\std{0.00}\\
	{0.040}\std{0.00}\\
	\textbf{0.035}\std{0.00}\\
	{0.040}\std{0.00}\\
	\textbf{0.035}\std{0.00}\\
	\end{tabular} &
	\begin{tabular}[c]{@{}c@{}}
	{0.011}\std{0.00}\\
	\textbf{0.026}\std{0.00}\\
	\textbf{0.026}\std{0.00}\\
	\textbf{0.026}\std{0.00}\\
	\textbf{0.026}\std{0.00}\\
	\textbf{0.026}\std{0.00}\\
	\textbf{0.026}\std{0.00}\\
	\end{tabular}\\
    \bottomrule
    \end{tabular}}
    \label{tab:german_faith_lr}
\end{table}

\begin{table}[h]
    \centering\small
    \renewcommand{\arraystretch}{0.9}
    \setlength{\tabcolsep}{1.5pt}
    \caption{
        \textbf{Ground-truth and predictive faithfulness results on the COMPAS dataset for all explanation methods with LR model.} Shown are average and standard error metric values computed across 1000 test instances. $\uparrow$ indicates that higher values are better, and $\downarrow$ indicates that lower values are better. Values corresponding to best performance are bolded.
    }
    {\begin{tabular}{lcccccccc}
    \toprule
    {{Method}} & {PRA~($\uparrow$)} & {RC~($\uparrow$)} & {FA~($\uparrow$)} & {RA~($\uparrow$)} & {SA~($\uparrow$)} & {SRA~($\uparrow$)} & {PGU~($\downarrow$)} & {PGI~($\uparrow$)} \\
    \toprule
    \begin{tabular}[l]{@{}l@{}}{Random}\\{VanillaGrad}\\{IntegratedGrad}\\{\GradtimesInput}\\{\SmoothGrad}\\{\SHAP}\\{\LIME}
    \end{tabular} &
	\begin{tabular}[c]{@{}c@{}}
	{0.504}\std{0.00}\\
	\textbf{1.000}\std{0.00}\\
	\textbf{1.000}\std{0.00}\\
	{0.622}\std{0.00}\\
	\textbf{1.000}\std{0.00}\\
	{0.614}\std{0.00}\\
	{0.977}\std{0.00}\\
	\end{tabular} &
	\begin{tabular}[c]{@{}c@{}}
	{0.009}\std{0.01}\\
	\textbf{1.000}\std{0.00}\\
	\textbf{1.000}\std{0.00}\\
	{0.320}\std{0.01}\\
	\textbf{1.000}\std{0.00}\\
	{0.287}\std{0.01}\\
	{0.981}\std{0.00}\\
	\end{tabular} &
	\begin{tabular}[c]{@{}c@{}}
	{0.209}\std{0.01}\\
	\textbf{1.000}\std{0.00}\\
	\textbf{1.000}\std{0.00}\\
	{0.311}\std{0.01}\\
	\textbf{1.000}\std{0.00}\\
	{0.311}\std{0.01}\\
	{0.999}\std{0.00}\\
	\end{tabular} &
	\begin{tabular}[c]{@{}c@{}}
	{0.137}\std{0.01}\\
	\textbf{1.000}\std{0.00}\\
	\textbf{1.000}\std{0.00}\\
	{0.142}\std{0.01}\\
	\textbf{1.000}\std{0.00}\\
	{0.134}\std{0.01}\\
	{0.997}\std{0.00}\\
	\end{tabular} &
	\begin{tabular}[c]{@{}c@{}}
	{0.107}\std{0.01}\\
	\textbf{1.000}\std{0.00}\\
	\textbf{1.000}\std{0.00}\\
	{0.311}\std{0.01}\\
	\textbf{1.000}\std{0.00}\\
	{0.310}\std{0.01}\\
	{0.999}\std{0.00}\\
	\end{tabular} &
	\begin{tabular}[c]{@{}c@{}}
	{0.067}\std{0.01}\\
	\textbf{1.000}\std{0.00}\\
	\textbf{1.000}\std{0.00}\\
	{0.142}\std{0.01}\\
	\textbf{1.000}\std{0.00}\\
	{0.134}\std{0.01}\\
	{0.997}\std{0.00}\\
	\end{tabular} &
	\begin{tabular}[c]{@{}c@{}}
	{0.072}\std{0.00}\\
	\textbf{0.055}\std{0.00}\\
	\textbf{0.055}\std{0.00}\\
	{0.070}\std{0.00}\\
	\textbf{0.055}\std{0.00}\\
	{0.069}\std{0.00}\\
	\textbf{0.055}\std{0.00}\\
	\end{tabular} &
	\begin{tabular}[c]{@{}c@{}}
	{0.029}\std{0.00}\\
	\textbf{0.058}\std{0.00}\\
	\textbf{0.058}\std{0.00}\\
	{0.033}\std{0.00}\\
	\textbf{0.058}\std{0.00}\\
	{0.033}\std{0.00}\\
	\textbf{0.058}\std{0.00}\\
	\end{tabular}\\
    \bottomrule
    \end{tabular}}
    \label{tab:compas_faith_lr}
\end{table}

\begin{table}[h]
    \centering\small
    \renewcommand{\arraystretch}{0.9}
    \setlength{\tabcolsep}{1.5pt}
    \caption{
        \textbf{Ground-truth and predictive faithfulness results on the GMSC dataset for all explanation methods with LR model.} Shown are average and standard error metric values computed across 1000 test instances. $\uparrow$ indicates that higher values are better, and $\downarrow$ indicates that lower values are better. Values corresponding to best performance are bolded.
    }
    {\begin{tabular}{lcccccccc}
    \toprule
    {{Method}} & {PRA~($\uparrow$)} & {RC~($\uparrow$)} & {FA~($\uparrow$)} & {RA~($\uparrow$)} & {SA~($\uparrow$)} & {SRA~($\uparrow$)} & {PGU~($\downarrow$)} & {PGI~($\uparrow$)} \\
    \toprule
    \begin{tabular}[l]{@{}l@{}}{Random}\\{VanillaGrad}\\{IntegratedGrad}\\{\GradtimesInput}\\{\SmoothGrad}\\{\SHAP}\\{\LIME}
    \end{tabular} &
	\begin{tabular}[c]{@{}c@{}}
	{0.499}\std{0.00}\\
	\textbf{1.000}\std{0.00}\\
	\textbf{1.000}\std{0.00}\\
	{0.499}\std{0.00}\\
	\textbf{1.000}\std{0.00}\\
	{0.495}\std{0.00}\\
	{0.909}\std{0.00}\\
	\end{tabular} &
	\begin{tabular}[c]{@{}c@{}}
	{-0.006}\std{0.01}\\
	\textbf{1.000}\std{0.00}\\
	\textbf{1.000}\std{0.00}\\
	{-0.006}\std{0.01}\\
	\textbf{1.000}\std{0.00}\\
	{-0.053}\std{0.01}\\
	{0.920}\std{0.00}\\
	\end{tabular} &
	\begin{tabular}[c]{@{}c@{}}
	{0.197}\std{0.01}\\
	\textbf{1.000}\std{0.00}\\
	\textbf{1.000}\std{0.00}\\
	{0.124}\std{0.01}\\
	\textbf{1.000}\std{0.00}\\
	{0.140}\std{0.01}\\
	{0.954}\std{0.00}\\
	\end{tabular} &
	\begin{tabular}[c]{@{}c@{}}
	{0.098}\std{0.01}\\
	\textbf{1.000}\std{0.00}\\
	\textbf{1.000}\std{0.00}\\
	{0.067}\std{0.01}\\
	\textbf{1.000}\std{0.00}\\
	{0.085}\std{0.01}\\
	{0.953}\std{0.00}\\
	\end{tabular} &
	\begin{tabular}[c]{@{}c@{}}
	{0.098}\std{0.01}\\
	\textbf{1.000}\std{0.00}\\
	\textbf{1.000}\std{0.00}\\
	{0.124}\std{0.01}\\
	\textbf{1.000}\std{0.00}\\
	{0.140}\std{0.01}\\
	{0.954}\std{0.00}\\
	\end{tabular} &
	\begin{tabular}[c]{@{}c@{}}
	{0.049}\std{0.00}\\
	\textbf{1.000}\std{0.00}\\
	\textbf{1.000}\std{0.00}\\
	{0.067}\std{0.01}\\
	\textbf{1.000}\std{0.00}\\
	{0.085}\std{0.01}\\
	{0.953}\std{0.00}\\
	\end{tabular} &
	\begin{tabular}[c]{@{}c@{}}
	{0.026}\std{0.00}\\
	\textbf{0.012}\std{0.00}\\
	\textbf{0.012}\std{0.00}\\
	{0.023}\std{0.00}\\
	\textbf{0.012}\std{0.00}\\
	{0.023}\std{0.00}\\
	\textbf{0.012}\std{0.00}\\
	\end{tabular} &
	\begin{tabular}[c]{@{}c@{}}
	{0.009}\std{0.00}\\
	\textbf{0.024}\std{0.00}\\
	\textbf{0.024}\std{0.00}\\
	{0.013}\std{0.00}\\
	\textbf{0.024}\std{0.00}\\
	{0.013}\std{0.00}\\
	\textbf{0.024}\std{0.00}\\
	\end{tabular}\\
    \bottomrule
    \end{tabular}}
    \label{tab:gmsc_faith_lr}
\end{table}

\begin{minipage}[c]{0.49\textwidth}
	\centering\small
    \renewcommand{\arraystretch}{0.9}
    \setlength{\tabcolsep}{2.0pt}
    \captionof{table}{
        \textbf{Stability results on the HELOC dataset for all explanation methods with LR model.} Shown are log average and standard error metric values computed across 1000 test instances. $\uparrow$ indicates that higher values are better, and $\downarrow$ indicates that lower values are better. Values corresponding to best performance are bolded.
    }
    {\begin{tabular}{lcc}
    \toprule
    {{Method}} & {RIS~($\downarrow$)} & {ROS~($\downarrow$)} \\
    \toprule
    \begin{tabular}[l]{@{}l@{}}{Random}\\{VanillaGrad}\\{IntegratedGrad}\\{\GradtimesInput}\\{\SmoothGrad}\\{\SHAP}\\{\LIME}
    \end{tabular} &
	\begin{tabular}[c]{@{}c@{}}
	{11.74}\std{0.00}\\
	{2.75}\std{0.02}\\
	\textbf{1.25}\std{0.01}\\
	{2.75}\std{0.02}\\
	{10.00}\std{0.03}\\
	{9.68}\std{0.02}\\
	{9.86}\std{0.03}\\
	\end{tabular} &
	\begin{tabular}[c]{@{}c@{}}
	{14.10}\std{0.02}\\
	{1.89}\std{0.13}\\
	\textbf{-0.42}\std{0.02}\\
	{3.64}\std{0.05}\\
	{12.27}\std{0.04}\\
	{12.01}\std{0.04}\\
	{12.82}\std{0.04}\\
	\end{tabular}\\
    \bottomrule
    \end{tabular}}
    \label{tab:heloc_stab_lr}
\end{minipage}
\quad
\begin{minipage}[c]{0.49\textwidth}
	\centering\small
    \renewcommand{\arraystretch}{0.9}
    \setlength{\tabcolsep}{2.0pt}
    \captionof{table}{
        \textbf{Stability results on the Adult Income dataset for all explanation methods with LR model.} Shown are log average and standard error metric values computed across 1000 test instances. $\uparrow$ indicates that higher values are better, and $\downarrow$ indicates that lower values are better. Values corresponding to best performance are bolded.
    }
    {\begin{tabular}{lcc}
    \toprule
    {{Method}} & {RIS~($\downarrow$)} & {ROS~($\downarrow$)} \\
    \toprule
    \begin{tabular}[l]{@{}l@{}}{Random}\\{VanillaGrad}\\{IntegratedGrad}\\{\GradtimesInput}\\{\SmoothGrad}\\{\SHAP}\\{\LIME}
    \end{tabular} &
	\begin{tabular}[c]{@{}c@{}}
	{12.95}\std{0.00}\\
	{3.75}\std{0.02}\\
	\textbf{2.61}\std{0.01}\\
	{3.79}\std{0.02}\\
	{11.41}\std{0.04}\\
	{11.36}\std{0.02}\\
	{11.35}\std{0.04}\\
	\end{tabular} &
	\begin{tabular}[c]{@{}c@{}}
	{14.43}\std{0.03}\\
	{2.34}\std{0.06}\\
	\textbf{0.23}\std{0.02}\\
	{3.46}\std{0.04}\\
	{13.33}\std{0.11}\\
	{12.12}\std{0.03}\\
	{14.23}\std{0.04}\\
	\end{tabular}\\
    \bottomrule
    \end{tabular}}
    \label{tab:adult_stab_lr}
\end{minipage}
\vspace{60pt}

\begin{minipage}[b]{0.48\textwidth}
	\centering\small
    \renewcommand{\arraystretch}{0.9}
    \setlength{\tabcolsep}{2.0pt}
    \captionof{table}{
        \textbf{Stability results on the COMPAS dataset for all explanation methods with LR model.} Shown are log average and standard error metric values computed across 1000 test instances. $\uparrow$ indicates that higher values are better, and $\downarrow$ indicates that lower values are better. Values corresponding to best performance are bolded.
    }
    {\begin{tabular}{lcc}
    \toprule
    {{Method}} & {RIS~($\downarrow$)} & {ROS~($\downarrow$)} \\
    \toprule
    \begin{tabular}[l]{@{}l@{}}{Random}\\{VanillaGrad}\\{IntegratedGrad}\\{\GradtimesInput}\\{\SmoothGrad}\\{\SHAP}\\{\LIME}
    \end{tabular} &
	\begin{tabular}[c]{@{}c@{}}
	{13.11}\std{0.01}\\
	{2.33}\std{0.02}\\
	\textbf{1.49}\std{0.01}\\
	{2.41}\std{0.01}\\
	{11.39}\std{0.02}\\
	{10.02}\std{0.02}\\
	{10.64}\std{0.02}\\
	\end{tabular} &
	\begin{tabular}[c]{@{}c@{}}
	{14.04}\std{0.02}\\
	{1.86}\std{0.15}\\
	\textbf{0.06}\std{0.02}\\
	{4.37}\std{0.05}\\
	{13.72}\std{0.04}\\
	{12.00}\std{0.03}\\
	{11.87}\std{0.04}\\
	\end{tabular}\\
    \bottomrule
    \end{tabular}}
    \label{tab:compas_stab_lr}
\end{minipage}
\hfill
\begin{minipage}[b]{0.48\textwidth}
	\centering\small
    \renewcommand{\arraystretch}{0.9}
    \setlength{\tabcolsep}{2.0pt}
    \captionof{table}{
        \textbf{Stability results on the GMSC dataset for all explanation methods with LR model.} Shown are log average and standard error metric values computed across 1000 test instances. $\uparrow$ indicates that higher values are better, and $\downarrow$ indicates that lower values are better. Values corresponding to best performance are bolded.
    }
    {\begin{tabular}{lcc}
    \toprule
    {{Method}} & {RIS~($\downarrow$)} & {ROS~($\downarrow$)} \\
    \toprule
    \begin{tabular}[l]{@{}l@{}}{Random}\\{VanillaGrad}\\{IntegratedGrad}\\{\GradtimesInput}\\{\SmoothGrad}\\{\SHAP}\\{\LIME}
    \end{tabular} &
	\begin{tabular}[c]{@{}c@{}}
	{11.47}\std{0.01}\\
	{2.77}\std{0.02}\\
	\textbf{1.87}\std{0.01}\\
	{2.84}\std{0.02}\\
	{10.52}\std{0.03}\\
	{9.25}\std{0.02}\\
	{10.44}\std{0.04}\\
	\end{tabular} &
	\begin{tabular}[c]{@{}c@{}}
	{14.97}\std{0.01}\\
	{2.86}\std{0.04}\\
	\textbf{1.27}\std{0.02}\\
	{4.79}\std{0.04}\\
	{11.97}\std{0.01}\\
	{13.18}\std{0.02}\\
	{13.48}\std{0.04}\\
	\end{tabular}\\
    \bottomrule
    \end{tabular}}
    \label{tab:gmsc_stab_lr}
\end{minipage}
\vspace{60pt}

\pagebreak
\subsection{Remaining Results on ANN models}
\label{app:results_ann}

\begin{minipage}[t]{0.48\textwidth}
	\centering\small
    \renewcommand{\arraystretch}{0.9}
    \setlength{\tabcolsep}{2.0pt}
    \captionof{table}{
        \textbf{Predictive faithfulness results on the Synthetic dataset for all explanation methods with ANN model.} Shown are average and standard error metric values computed across 1000 test instances. $\uparrow$ indicates that higher values are better, and $\downarrow$ indicates that lower values are better. Values corresponding to best performance are bolded.
    }
    {\begin{tabular}{lcc}
    \toprule
    {{Method}} & {PGU~($\downarrow$)} & {PGI~($\uparrow$)} \\
    \toprule
    \begin{tabular}[l]{@{}l@{}}{Random}\\{VanillaGrad}\\{IntegratedGrad}\\{\GradtimesInput}\\{\SmoothGrad}\\{\SHAP}\\{\LIME}
    \end{tabular} &
	\begin{tabular}[c]{@{}c@{}}
	{0.143}\std{0.005}\\
	{0.107}\std{0.004}\\
	{0.116}\std{0.004}\\
	{0.113}\std{0.004}\\
	{0.106}\std{0.004}\\
	{0.124}\std{0.005}\\
	\textbf{0.105}\std{0.004}\\
	\end{tabular} &
	\begin{tabular}[c]{@{}c@{}}
	{0.053}\std{0.003}\\
	{0.116}\std{0.004}\\
	{0.102}\std{0.004}\\
	{0.109}\std{0.004}\\
	{0.116}\std{0.004}\\
	{0.092}\std{0.004}\\
	\textbf{0.117}\std{0.004}\\
	\end{tabular}\\
    \bottomrule
    \end{tabular}}
    \label{tab:gaussian_faith_ann}
\end{minipage}
\hfill
\begin{minipage}[t]{0.48\textwidth}
	\centering\small
    \renewcommand{\arraystretch}{0.9}
    \setlength{\tabcolsep}{2.0pt}
    \captionof{table}{
        \textbf{Predictive faithfulness results on the German Credit dataset for all explanation methods with ANN model.} Shown are average and standard error metric values computed across 1000 test instances. $\uparrow$ indicates that higher values are better, and $\downarrow$ indicates that lower values are better. Values corresponding to best performance are bolded.
    }
    {\begin{tabular}{lcc}
    \toprule
    {{Method}} & {PGU~($\downarrow$)} & {PGI~($\uparrow$)} \\
    \toprule
    \begin{tabular}[l]{@{}l@{}}{Random}\\{VanillaGrad}\\{IntegratedGrad}\\{\GradtimesInput}\\{\SmoothGrad}\\{\SHAP}\\{\LIME}
    \end{tabular} &
	\begin{tabular}[c]{@{}c@{}}
	{0.120}\std{0.007}\\
	{0.104}\std{0.007}\\
	{0.105}\std{0.007}\\
	\textbf{0.103}\std{0.007}\\
	\textbf{0.103}\std{0.007}\\
	\textbf{0.103}\std{0.007}\\
	{0.104}\std{0.007}\\
	\end{tabular} &
	\begin{tabular}[c]{@{}c@{}}
	{0.030}\std{0.002}\\
	{0.056}\std{0.004}\\
	{0.053}\std{0.004}\\
	\textbf{0.065}\std{0.004}\\
	{0.057}\std{0.004}\\
	{0.063}\std{0.004}\\
	{0.056}\std{0.004}\\
	\end{tabular}\\
    \bottomrule
    \end{tabular}}
    \label{tab:german_faith_ann}
\end{minipage}
\vspace{50pt}

  \begin{minipage}[t]{0.48\textwidth}
	\centering\small
    \renewcommand{\arraystretch}{0.9}
    \setlength{\tabcolsep}{2.0pt}
    \captionof{table}{
        \textbf{Predictive faithfulness results on the HELOC dataset for all explanation methods with ANN model.} Shown are average and standard error metric values computed across 1000 test instances. $\uparrow$ indicates that higher values are better, and $\downarrow$ indicates that lower values are better. Values corresponding to best performance are bolded.
    }
    {\begin{tabular}{lcc}
    \toprule
    {{Method}} & {PGU~($\downarrow$)} & {PGI~($\uparrow$)} \\
    \toprule
    \begin{tabular}[l]{@{}l@{}}{Random}\\{VanillaGrad}\\{IntegratedGrad}\\{\GradtimesInput}\\{\SmoothGrad}\\{\SHAP}\\{\LIME}
    \end{tabular} &
	\begin{tabular}[c]{@{}c@{}}
	{0.107}\std{0.003}\\
	\textbf{0.082}\std{0.003}\\
	{0.084}\std{0.003}\\
	{0.106}\std{0.003}\\
	\textbf{0.082}\std{0.003}\\
	{0.106}\std{0.003}\\
	\textbf{0.082}\std{0.003}\\
	\end{tabular} &
	\begin{tabular}[c]{@{}c@{}}
	{0.042}\std{0.002}\\
	{0.081}\std{0.003}\\
	{0.079}\std{0.003}\\
	{0.046}\std{0.002}\\
	\textbf{0.082}\std{0.003}\\
	{0.044}\std{0.002}\\
	{0.081}\std{0.003}\\
	\end{tabular}\\
    \bottomrule
    \end{tabular}}
    \label{tab:heloc_faith_ann}
  \end{minipage}
  \hfill
  \begin{minipage}[t]{0.48\textwidth}
	\centering\small
    \renewcommand{\arraystretch}{0.9}
    \setlength{\tabcolsep}{2.0pt}
    \captionof{table}{
        \textbf{Predictive faithfulness results on the Adult Income dataset for all explanation methods with ANN model.} Shown are average and standard error metric values computed across 1000 test instances. $\uparrow$ indicates that higher values are better, and $\downarrow$ indicates that lower values are better. Values corresponding to best performance are bolded.
    }
    {\begin{tabular}{lcc}
    \toprule
    {{Method}} & {PGU~($\downarrow$)} & {PGI~($\uparrow$)} \\
    \toprule
    \begin{tabular}[l]{@{}l@{}}{Random}\\{VanillaGrad}\\{IntegratedGrad}\\{\GradtimesInput}\\{\SmoothGrad}\\{\SHAP}\\{\LIME}
    \end{tabular} &
	\begin{tabular}[c]{@{}c@{}}
	{0.207}\std{0.004}\\
	{0.071}\std{0.003}\\
	\textbf{0.070}\std{0.003}\\
	{0.221}\std{0.004}\\
	\textbf{0.070}\std{0.003}\\
	{0.223}\std{0.004}\\
	{0.071}\std{0.003}\\
	\end{tabular} &
	\begin{tabular}[c]{@{}c@{}}
	{0.069}\std{0.003}\\
	{0.230}\std{0.004}\\
	\textbf{0.233}\std{0.004}\\
	{0.071}\std{0.003}\\
	{0.230}\std{0.004}\\
	{0.068}\std{0.003}\\
	{0.230}\std{0.004}\\
	\end{tabular}\\
    \bottomrule
    \end{tabular}}
    \label{tab:adult_faith_ann}
  \end{minipage}
\vspace{50pt}

  \begin{minipage}[t]{0.48\textwidth}
	\centering\small
    \renewcommand{\arraystretch}{0.9}
    \setlength{\tabcolsep}{2.0pt}
    \captionof{table}{
        \textbf{Predictive faithfulness results on the COMPAS dataset for all explanation methods with ANN model.} Shown are average and standard error metric values computed across 1000 test instances. $\uparrow$ indicates that higher values are better, and $\downarrow$ indicates that lower values are better. Values corresponding to best performance are bolded.
    }
    {\begin{tabular}{lcc}
    \toprule
    {{Method}} & {PGU~($\downarrow$)} & {PGI~($\uparrow$)} \\
    \toprule
    \begin{tabular}[l]{@{}l@{}}{Random}\\{VanillaGrad}\\{IntegratedGrad}\\{\GradtimesInput}\\{\SmoothGrad}\\{\SHAP}\\{\LIME}
    \end{tabular} &
	\begin{tabular}[c]{@{}c@{}}
	{0.095}\std{0.004}\\
	\textbf{0.056}\std{0.003}\\
	\textbf{0.056}\std{0.003}\\
	{0.073}\std{0.003}\\
	\textbf{0.056}\std{0.003}\\
	{0.073}\std{0.003}\\
	{0.057}\std{0.003}\\
	\end{tabular} &
	\begin{tabular}[c]{@{}c@{}}
	{0.035}\std{0.002}\\
	\textbf{0.090}\std{0.004}\\
	\textbf{0.090}\std{0.004}\\
	{0.068}\std{0.003}\\
	\textbf{0.090}\std{0.004}\\
	{0.072}\std{0.003}\\
	\textbf{0.090}\std{0.004}\\
	\end{tabular}\\
    \bottomrule
    \end{tabular}}
    \label{tab:compas_faith_ann}
  \end{minipage}
  \hfill
  \begin{minipage}[t]{0.48\textwidth}
	\centering\small
    \renewcommand{\arraystretch}{0.9}
    \setlength{\tabcolsep}{2.0pt}
    \captionof{table}{
        \textbf{Predictive faithfulness results on the GMSC dataset for all explanation methods with ANN model.} Shown are average and standard error metric values computed across 1000 test instances. $\uparrow$ indicates that higher values are better, and $\downarrow$ indicates that lower values are better. Values corresponding to best performance are bolded.
    }
    {\begin{tabular}{lcc}
    \toprule
    {{Method}} & {PGU~($\downarrow$)} & {PGI~($\uparrow$)} \\
    \toprule
    \begin{tabular}[l]{@{}l@{}}{Random}\\{VanillaGrad}\\{IntegratedGrad}\\{\GradtimesInput}\\{\SmoothGrad}\\{\SHAP}\\{\LIME}
    \end{tabular} &
	\begin{tabular}[c]{@{}c@{}}
	{0.088}\std{0.003}\\
	\textbf{0.015}\std{0.001}\\
	\textbf{0.015}\std{0.001}\\
	{0.059}\std{0.002}\\
	\textbf{0.015}\std{0.001}\\
	{0.067}\std{0.002}\\
	\textbf{0.015}\std{0.001}\\
	\end{tabular} &
	\begin{tabular}[c]{@{}c@{}}
	{0.026}\std{0.002}\\
	\textbf{0.102}\std{0.003}\\
	\textbf{0.102}\std{0.003}\\
	{0.063}\std{0.003}\\
	\textbf{0.102}\std{0.003}\\
	{0.053}\std{0.003}\\
	{0.101}\std{0.003}\\
	\end{tabular}\\
    \bottomrule
    \end{tabular}}
    \label{tab:gmsc_faith_ann}
  \end{minipage}
\vspace{50pt}

\pagebreak

  \begin{minipage}[b]{0.48\textwidth}
	\centering\small
    \renewcommand{\arraystretch}{0.9}
    \setlength{\tabcolsep}{2.0pt}
    \captionof{table}{
        \textbf{Stability results on the Synthetic dataset for all explanation methods with ANN model.} Shown are log average and standard error metric values computed across 1000 test instances. $\uparrow$ indicates that higher values are better, and $\downarrow$ indicates that lower values are better. Values corresponding to best performance are bolded.
    }
    {\begin{tabular}{lccc}
    \toprule
    {{Method}} & {RIS~($\downarrow$)} & {RRS~($\downarrow$)} & {ROS~($\downarrow$)} \\
    \toprule
    \begin{tabular}[l]{@{}l@{}}{Random}\\{VanillaGrad}\\{IntegratedGrad}\\{\GradtimesInput}\\{\SmoothGrad}\\{\SHAP}\\{\LIME}
    \end{tabular} &
	\begin{tabular}[c]{@{}c@{}}
	{11.58}\std{0.00}\\
	{5.29}\std{0.20}\\
	\textbf{3.56}\std{0.17}\\
	{5.21}\std{0.18}\\
	{31.27}\std{1.00}\\
	{10.55}\std{0.02}\\
	{21.11}\std{0.89}\\
	\end{tabular} &
	\begin{tabular}[c]{@{}c@{}}
	{10.80}\std{0.01}\\
	{4.23}\std{0.22}\\
	\textbf{2.51}\std{0.20}\\
	{4.13}\std{0.20}\\
	{30.62}\std{1.00}\\
	{9.74}\std{0.02}\\
	{20.60}\std{0.91}\\
	\end{tabular} &
	\begin{tabular}[c]{@{}c@{}}
	{15.12}\std{0.04}\\
	{5.47}\std{0.48}\\
	{5.79}\std{0.39}\\
	\textbf{5.33}\std{0.37}\\
	{32.06}\std{1.00}\\
	{13.94}\std{0.04}\\
	{21.70}\std{0.82}\\
	\end{tabular}\\
    \bottomrule
    \end{tabular}}
    \label{tab:gaussian_stab_ann}
  \end{minipage}
  \hfill
  \begin{minipage}[b]{0.48\textwidth}
	\centering\small
    \renewcommand{\arraystretch}{0.9}
    \setlength{\tabcolsep}{2.0pt}
    \captionof{table}{
        \textbf{Stability results on the German Credit dataset for all explanation methods with ANN model.} Shown are log average and standard error metric values computed across 1000 test instances. $\uparrow$ indicates that higher values are better, and $\downarrow$ indicates that lower values are better. Values corresponding to best performance are bolded.
    }
    {\begin{tabular}{lccc}
    \toprule
    {{Method}} & {RIS~($\downarrow$)} & {RRS~($\downarrow$)} & {ROS~($\downarrow$)} \\
    \toprule
    \begin{tabular}[l]{@{}l@{}}{Random}\\{VanillaGrad}\\{IntegratedGrad}\\{\GradtimesInput}\\{\SmoothGrad}\\{\SHAP}\\{\LIME}
    \end{tabular} &
	\begin{tabular}[c]{@{}c@{}}
	{12.84}\std{0.00}\\
	{3.18}\std{0.31}\\
	\textbf{2.16}\std{0.33}\\
	{3.19}\std{0.31}\\
	{11.57}\std{0.06}\\
	{11.80}\std{0.01}\\
	{11.77}\std{0.05}\\
	\end{tabular} &
	\begin{tabular}[c]{@{}c@{}}
	{12.77}\std{0.01}\\
	{3.06}\std{0.36}\\
	\textbf{1.98}\std{0.34}\\
	{3.06}\std{0.36}\\
	{11.47}\std{0.06}\\
	{11.77}\std{0.02}\\
	{11.73}\std{0.05}\\
	\end{tabular} &
	\begin{tabular}[c]{@{}c@{}}
	{16.06}\std{0.06}\\
	\textbf{3.24}\std{0.43}\\
	{4.34}\std{0.36}\\
	{3.96}\std{0.21}\\
	{14.40}\std{0.11}\\
	{15.13}\std{0.07}\\
	{14.78}\std{0.10}\\
	\end{tabular}\\
    \bottomrule
    \end{tabular}}
    \label{tab:german_stab_ann}
  \end{minipage}
\vspace{50pt}

  \begin{minipage}[b]{0.48\textwidth}
	\centering\small
    \renewcommand{\arraystretch}{0.9}
    \setlength{\tabcolsep}{2.0pt}
    \captionof{table}{
        \textbf{Stability results on the HELOC dataset for all explanation methods with ANN model.} Shown are log average and standard error metric values computed across 1000 test instances. $\uparrow$ indicates that higher values are better, and $\downarrow$ indicates that lower values are better. Values corresponding to best performance are bolded.
    }
    {\begin{tabular}{lccc}
    \toprule
    {{Method}} & {RIS~($\downarrow$)} & {RRS~($\downarrow$)} & {ROS~($\downarrow$)} \\
    \toprule
    \begin{tabular}[l]{@{}l@{}}{Random}\\{VanillaGrad}\\{IntegratedGrad}\\{\GradtimesInput}\\{\SmoothGrad}\\{\SHAP}\\{\LIME}
    \end{tabular} &
	\begin{tabular}[c]{@{}c@{}}
	{11.74}\std{0.00}\\
	{3.53}\std{0.17}\\
	\textbf{2.47}\std{0.07}\\
	{3.69}\std{0.23}\\
	{10.96}\std{0.08}\\
	{10.69}\std{0.02}\\
	{10.42}\std{0.05}\\
	\end{tabular} &
	\begin{tabular}[c]{@{}c@{}}
	{11.52}\std{0.01}\\
	{3.00}\std{0.23}\\
	\textbf{2.03}\std{0.09}\\
	{3.18}\std{0.29}\\
	{10.85}\std{0.07}\\
	{10.54}\std{0.02}\\
	{10.32}\std{0.05}\\
	\end{tabular} &
	\begin{tabular}[c]{@{}c@{}}
	{15.16}\std{0.04}\\
	{4.49}\std{0.50}\\
	\textbf{4.10}\std{0.18}\\
	{5.32}\std{0.33}\\
	{13.91}\std{0.06}\\
	{14.32}\std{0.04}\\
	{14.02}\std{0.08}\\
	\end{tabular}\\
    \bottomrule
    \end{tabular}}
    \label{tab:heloc_stab_ann}
    \end{minipage}
  \hfill
  \begin{minipage}[b]{0.48\textwidth}
	\centering\small
    \renewcommand{\arraystretch}{0.9}
    \setlength{\tabcolsep}{2.0pt}
    \captionof{table}{
        \textbf{Stability results on the Adult Income dataset for all explanation methods with ANN model.} Shown are log average and standard error metric values computed across 1000 test instances. $\uparrow$ indicates that higher values are better, and $\downarrow$ indicates that lower values are better. Values corresponding to best performance are bolded.
    }
    {\begin{tabular}{lccc}
    \toprule
    {{Method}} & {RIS~($\downarrow$)} & {RRS~($\downarrow$)} & {ROS~($\downarrow$)} \\
    \toprule
    \begin{tabular}[l]{@{}l@{}}{Random}\\{VanillaGrad}\\{IntegratedGrad}\\{\GradtimesInput}\\{\SmoothGrad}\\{\SHAP}\\{\LIME}
    \end{tabular} &
	\begin{tabular}[c]{@{}c@{}}
	{12.95}\std{0.00}\\
	{5.73}\std{0.07}\\
	\textbf{3.81}\std{0.11}\\
	{5.79}\std{0.10}\\
	{13.33}\std{0.31}\\
	{13.31}\std{0.09}\\
	{10.83}\std{0.01}\\
	\end{tabular} &
	\begin{tabular}[c]{@{}c@{}}
	{12.03}\std{0.01}\\
	{3.98}\std{0.16}\\
	\textbf{1.97}\std{0.13}\\
	{4.02}\std{0.15}\\
	{13.98}\std{0.39}\\
	{12.01}\std{0.09}\\
	{9.67}\std{0.02}\\
	\end{tabular} &
	\begin{tabular}[c]{@{}c@{}}
	{14.75}\std{0.05}\\
	{3.58}\std{0.33}\\
	\textbf{3.38}\std{0.07}\\
	{4.20}\std{0.17}\\
	{17.08}\std{0.45}\\
	{14.49}\std{0.12}\\
	{12.95}\std{0.06}\\
	\end{tabular}\\
    \bottomrule
    \end{tabular}}
    \label{tab:adult_stab_ann}
    \end{minipage}
\vspace{50pt}

  \begin{minipage}[b]{0.48\textwidth}
	\centering\small
    \renewcommand{\arraystretch}{0.9}
    \setlength{\tabcolsep}{2.0pt}
    \captionof{table}{
        \textbf{Stability results on the COMPAS dataset for all explanation methods with ANN model.} Shown are log average and standard error metric values computed across 1000 test instances. $\uparrow$ indicates that higher values are better, and $\downarrow$ indicates that lower values are better. Values corresponding to best performance are bolded.
    }
    {\begin{tabular}{lccc}
    \toprule
    {{Method}} & {RIS~($\downarrow$)} & {RRS~($\downarrow$)} & {ROS~($\downarrow$)} \\
    \toprule
    \begin{tabular}[l]{@{}l@{}}{Random}\\{VanillaGrad}\\{IntegratedGrad}\\{\GradtimesInput}\\{\SmoothGrad}\\{\SHAP}\\{\LIME}
    \end{tabular} &
	\begin{tabular}[c]{@{}c@{}}
	{13.11}\std{0.01}\\
	{4.68}\std{0.18}\\
	\textbf{3.59}\std{0.04}\\
	{4.99}\std{0.30}\\
	{17.33}\std{0.13}\\
	{10.55}\std{0.02}\\
	{15.23}\std{0.10}\\
	\end{tabular} &
	\begin{tabular}[c]{@{}c@{}}
	{12.88}\std{0.01}\\
	{3.96}\std{0.19}\\
	\textbf{2.83}\std{0.05}\\
	{4.30}\std{0.32}\\
	{17.21}\std{0.12}\\
	{10.39}\std{0.01}\\
	{15.13}\std{0.09}\\
	\end{tabular} &
	\begin{tabular}[c]{@{}c@{}}
	{15.30}\std{0.03}\\
	{4.30}\std{0.53}\\
	\textbf{3.86}\std{0.30}\\
	{5.46}\std{0.42}\\
	{18.09}\std{0.10}\\
	{12.80}\std{0.04}\\
	{16.37}\std{0.11}\\
	\end{tabular}\\
    \bottomrule
    \end{tabular}}
    \label{tab:compas_stab_ann}
  \end{minipage}
  \hfill
  \begin{minipage}[b]{0.48\textwidth}
	\centering\small
    \renewcommand{\arraystretch}{0.9}
    \setlength{\tabcolsep}{2.0pt}
    \captionof{table}{
        \textbf{Stability results on the GMSC dataset for all explanation methods with ANN model.} Shown are log average and standard error metric values computed across 1000 test instances. $\uparrow$ indicates that higher values are better, and $\downarrow$ indicates that lower values are better. Values corresponding to best performance are bolded.
    }
    {\begin{tabular}{lccc}
    \toprule
    {{Method}} & {RIS~($\downarrow$)} & {RRS~($\downarrow$)} & {ROS~($\downarrow$)} \\
    \toprule
    \begin{tabular}[l]{@{}l@{}}{Random}\\{VanillaGrad}\\{IntegratedGrad}\\{\GradtimesInput}\\{\SmoothGrad}\\{\SHAP}\\{\LIME}
    \end{tabular} &
	\begin{tabular}[c]{@{}c@{}}
	{11.47}\std{0.01}\\
	{4.44}\std{0.09}\\
	\textbf{3.63}\std{0.02}\\
	{4.59}\std{0.11}\\
	{10.35}\std{0.03}\\
	{9.94}\std{0.02}\\
	{9.77}\std{0.03}\\
	\end{tabular} &
	\begin{tabular}[c]{@{}c@{}}
	{11.51}\std{0.01}\\
	{3.78}\std{0.11}\\
	\textbf{2.94}\std{0.03}\\
	{3.96}\std{0.14}\\
	{10.28}\std{0.03}\\
	{9.94}\std{0.01}\\
	{9.76}\std{0.03}\\
	\end{tabular} &
	\begin{tabular}[c]{@{}c@{}}
	{14.72}\std{0.04}\\
	\textbf{4.22}\std{0.55}\\
	{4.40}\std{0.26}\\
	{5.57}\std{0.48}\\
	{13.29}\std{0.05}\\
	{12.73}\std{0.06}\\
	{13.65}\std{0.05}\\
	\end{tabular}\\
    \bottomrule
    \end{tabular}}
    \label{tab:gmsc_stab_ann}
  \end{minipage}
\vspace{50pt}

\begin{minipage}{\textwidth}
    \begin{minipage}[b]{0.48\textwidth}
    \centering
    \includegraphics[width=0.99\textwidth,center]{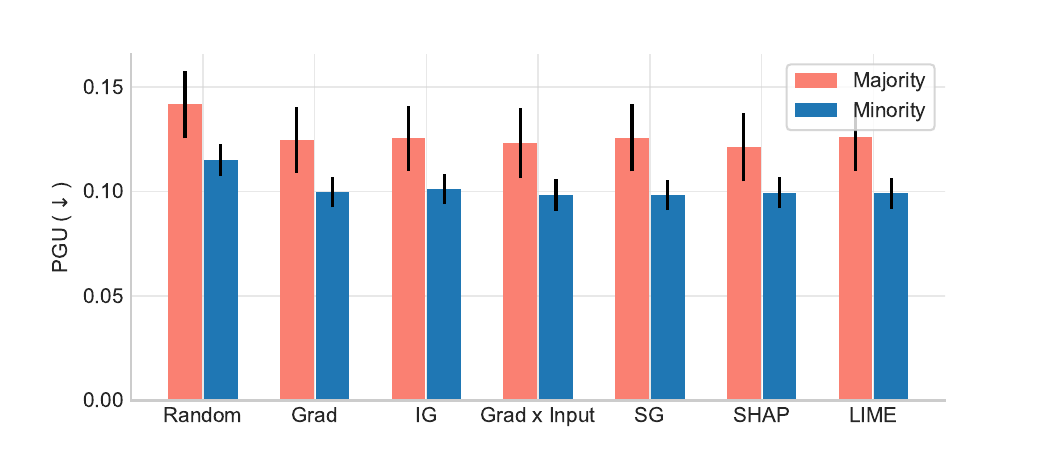}
    \captionof{figure}{
        \textbf{Fairness analysis of PGU metric on the German Credit dataset with ANN model.} Shown are average and standard error values for majority (male) and minority (female) subgroups. Larger gaps between the values of majority and minority subgroups (i.e, red and blue bars respectively) indicate higher disparities which are undesirable.
    }
    \label{fig:german_disparity_ann}
  \end{minipage}
  \hfill
  \begin{minipage}[b]{0.48\textwidth}
    \centering
    \includegraphics[width=0.99\textwidth,center]{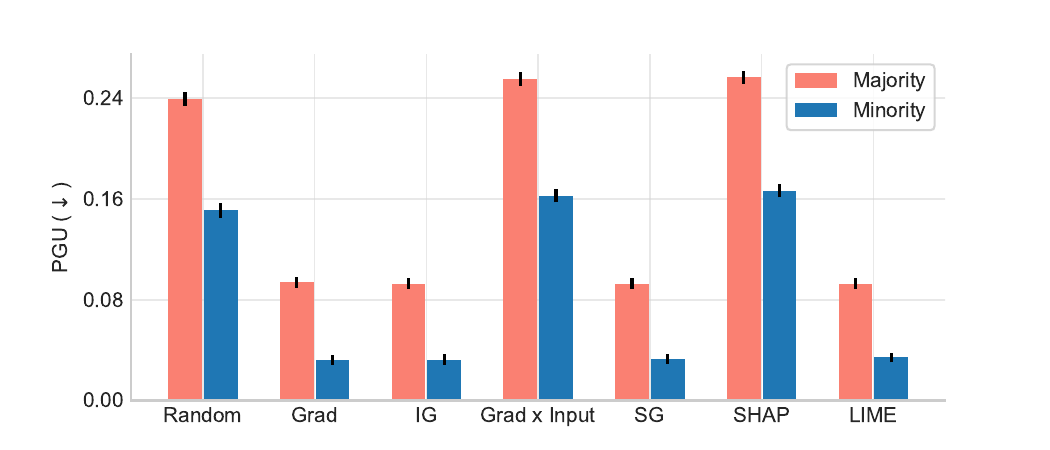}
    \captionof{figure}{
        \textbf{Fairness analysis of PGU metric on the Adult Income dataset with ANN model.} Shown are average and standard error values for majority (male) and minority (female) subgroups. Larger gaps between the values of majority and minority subgroups (i.e, red and blue bars respectively) indicate higher disparities which are undesirable.
    }
    \label{fig:adult_disparity_ann}
  \end{minipage}
\end{minipage}

\section{Choice of XAI methods, datasets, and models}
\label{app:choice}
While feature attribution-based explanation methods such as LIME, SHAP, and Gradient-based methods have been proposed a few years back, they continue to be the most popular and widely used post hoc explanation methods both in research~\citep{chen2022usecase,krishna2022disagreement,alvarez2018robustness,hooker2018evaluating,han2022explanation} and in practice~\citep{elshawi2019interpretability,whitmore2016mapping,ibrahim2019global,ghassemi2021false}. In fact, several recent works published in 2022 have analyzed these methods both theoretically and empirically, and have called for further study of these methods given their widespread adoption~\citep{dasgupta2022framework,chen2022usecase,balagopalan2022road,han2022explanation,krishna2022disagreement,fokkema2022attribution}. Furthermore, recent research has also argued that there is little to no understanding of the behavior and effectiveness of even basic post hoc explanation methods such as LIME, SHAP, and gradient-based methods~\citep{krishna2022disagreement,lipton2016mythos,rudin2019stop,kaur2020interpreting,bansal2020sam,agarwal2020explaining}, and developing such an understanding would be a critical first step towards the progress of the XAI field. To this end, we focus on these methods for the first release of our OpenXAI framework. In the next release, we plan to evaluate and benchmark other recently proposed methods (e.g., TCAV and its extensions, influence functions, etc.) as well.

Note that the evaluation metrics and the explanation methods that we include in our framework are generic enough to be applicable to other modalities of data including text and images. The reason why we focused on tabular data for the first release of OpenXAI is two-fold: i) The need for model understanding and explainability is often motivated by high-stakes decision-making settings and applications e.g., loan approvals, disease diagnosis and treatment recommendations, recidivism prediction etc.~\citep{borisov2021deep,pawelczyk2020learning,pawelczyk2021carla}. Data encountered in these settings is predominantly tabular. ii) Recent research has argued that there is no clear understanding as to which explanation methods perform well on what kinds of metrics even on simple, low-dimensional tabular datasets~\citep{lipton2016mythos,dasgupta2022framework,krishna2022disagreement,han2022explanation}, and that this is a big open question which has far reaching implications for the progress of the field. To this end, we focused on tabular data for the first release of OpenXAI so that we could find some answers to the aforementioned question in the context of simple, low-dimensional tabular datasets before proceeding to high-dimensional image and text datasets. In the next release of OpenXAI, we plan to include and support image and text datasets, and also add metrics and explanation methods that are specific to these new data modalities.

The datasets that we utilize in this work are very popular and are widely employed in XAI and fairness research till date. For instance, several recent works in XAI published at ICML, NeurIPS, and FAccT conferences in 2021-22 have employed these datasets both to evaluate the efficacy of newly proposed methods, as well as to study the behavior of existing methods~\citep{balagopalan2022road,dai2022fairness,dasgupta2022framework,dominguez2022adversarial,karimi2020algorithmic,slack2021reliable,upadhyay2021towards,agarwal2021towards}. Given this, we follow suit and employ these datasets in our benchmarking efforts. Similarly, the aforementioned works also employ logistic regression models and deep neural network architectures similar to the ones considered in our research.


\end{document}